\documentclass{article} \pdfminorversion=6

    \PassOptionsToPackage{numbers, compress}{natbib}
 \usepackage[preprint]{neurips_2026}

\usepackage[utf8]{inputenc} % allow utf-8 input
\usepackage[T1]{fontenc}    % use 8-bit T1 fonts
\usepackage{hyperref}       % hyperlinks
\usepackage{url}            % simple URL typesetting
\usepackage{booktabs}       % professional-quality tables
\usepackage{amsfonts}       % blackboard math symbols
\usepackage{nicefrac}       % compact symbols for 1/2, etc.
\usepackage{microtype}      % microtypography
\usepackage{xcolor}         % colors
\usepackage{amsmath, amssymb, amsthm}

\usepackage{graphicx}
\usepackage{wrapfig} 
\usepackage[dvipsnames]{xcolor}
\usepackage{caption}
\usepackage{algorithm}    
\usepackage{enumitem}
\usepackage{algpseudocode}                                               \usepackage{makecell}

\title{Which Tokens to Merge? Diffusion Dynamics\\for Efficient Image Generation}

\author{
SeungJu Cha \quad
Ye-Chan Kim \quad
HyunGee Kim \quad
Sungho Koh \quad
Dong-Jin Kim\textsuperscript{$\dagger$}
\\[0.5em]
Hanyang University, South Korea
\\
\texttt{\{sju9020, dpcksdl78, khjiiii2002, ksh000906, djdkim\}@hanyang.ac.kr}
\\
{\footnotesize $\dagger$ Corresponding author}
}

\begin{document}

\maketitle

\begin{abstract}
Token merging has emerged as an effective training-free strategy 
for accelerating diffusion models by reducing redundant token 
computation while minimizing generation degradation. A key 
challenge is determining which tokens can be safely merged and 
which should be preserved for subsequent computation. 
Existing method addresses this by preserving prompt-aligned foreground regions, thereby protecting semantically relevant structures. 
However, this text-relevance criterion overlooks background objects, local details that are not explicitly mentioned in the prompt, leading to degraded structural coherence.
In this work, we propose \textbf{D}ynamics-\textbf{a}ware \textbf{T}oken \textbf{Me}rging \textbf{(DaTMe)}, 
a training-free framework that grounds token merging in the 
underlying diffusion dynamics. 
Our key insight is that token importance should reflect whether a token remains unresolved during denoising and whether it carries structural information. 
To this end, DaTMe constructs a dynamics-aware importance map from the Tweedie denoised 
estimate. 
The map combines a temporal signal, measuring inter-step changes in the predicted clean image, with a structural signal, capturing local high-frequency content such as boundaries and fine details. 
Using this map, DaTMe adaptively assigns tokens into merging roles under a given compression ratio: important tokens are preserved as anchors, while tokens that are both 
temporally stabilized and structurally homogeneous are selected as safe merge candidates.
Experiments on PixArt-$\alpha$ and FLUX demonstrate that DaTMe maintains visual fidelity across the entire image, highlighting diffusion dynamics as an effective token-level merging decision in efficient generation. Our code is available at \url{https://github.com/SeungJuCha/DaTMe}.
\end{abstract}

\section{Introduction}
\label{sec:intro}

Diffusion models~\cite{rombach2022high, peebles2023scalable} have become a dominant paradigm for high-quality image~\cite{podell2023sdxl, esser2024scaling, chen2023pixartalphafasttrainingdiffusion} and video synthesis~\cite{li2024t2v, kong2024hunyuanvideo, polyak2024movie}, owing to their strong fidelity and scalability.
However, their iterative denoising process incurs substantial computational cost, as a large number of tokens are repeatedly processed at every inference step~\cite{Wu_2025_ICCV}.
To address this, token merging~\cite{bolya2023token} has emerged as an effective training-free acceleration strategy that groups similar tokens, merges them into representative tokens, and uses the merged representations in subsequent layers~\cite{Wu_2025_ICCV,bolya2023token, zhang2025training, lu2025toma, fang2025attend}.
For example, early token merging methods such as ToMeSD~\cite{bolya2023token} first construct local merge candidate groups in a content-agnostic manner, and then form merge pairs simply based on feature similarity within each group.
Consequently, content-relevant tokens can be selected as merge sources, causing local details to be absorbed and leading to visual distortion~\cite{Wu_2025_ICCV}.
% This raises a central challenge: \textit{which important tokens should be preserved}\djkim{(important?)} at each timestep to maintain image quality under token merging.
This raises a central challenge: \emph{which tokens should be absorbed through merging}, and \emph{which must be retained} as anchors for subsequent computation.

\begin{figure}[t]
    \centering
    \includegraphics[width=1.0\linewidth]{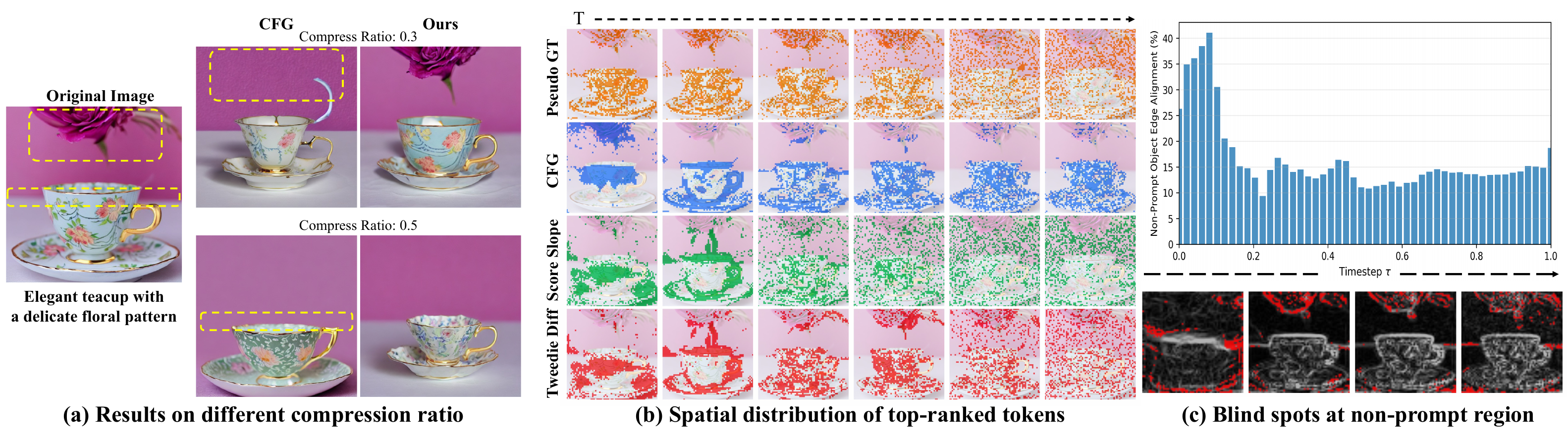}
 \caption{
\textbf{(a) Comparison under different compression ratios.}
CFG-based merging~\cite{Wu_2025_ICCV} often overlooks background details and structures, whereas our method better preserves background integrity.
\textbf{(b) Spatial distribution of top-ranked tokens over denoising timesteps.}
CFG-based importance remains concentrated on the foreground object with limited temporal variation. 
In contrast, our score-based signals (Score Slope and Tweedie Diff) progressively shift from object regions to fine-grained spatial details, better matching the pseudo-GT pattern.
Here, pseudo-GT is defined as the difference between the final denoised latent and the Tweedie estimate at each timestep.
\textbf{(c) Non-prompt region blind spot.} Low-ranked tokens under the CFG-based signal consistently overlap with structurally salient regions in text-irrelevant areas (red).
}
    
    \label{fig:problem_def}
\end{figure}

Prior work~\cite{Wu_2025_ICCV} leverages classifier-free guidance (CFG)~\cite{ho2022classifier} as a token importance signal, identifying tokens strongly steered by the text condition through the discrepancy between conditional and unconditional noise predictions as anchors to be retained.
However, as shown in Fig.~\ref{fig:problem_def}(a), CFG-based 
merging tends to underprotect non-prompt regions of the image, losing 
fine details and structural coherence. 
We attribute this to CFG's prompt-centric bias: by 
focusing exclusively on text-conditioned influence, CFG overlooks two key aspects of how diffusion models generate images.

First, CFG fails to capture \textbf{time-varying token importance}.
Diffusion generation follows a coarse-to-fine progression~\cite{hertz2022prompt, hwang2023resolution},
where tokens undergoing large inter-step changes represent regions in which the model is actively injecting details~\cite{cao2025temporal}.
% making the model's focus highly time-dependent as it shifts from object-level structures to finer details across the image.
In this context, prematurely merging such still-evolving tokens directly attenuates their ongoing refinement by averaging their features with neighboring tokens.
Crucially, the spatial distribution of these actively evolving tokens is time-varying: while it remains concentrated on foreground objects in early steps, it progressively expands toward broader spatial regions, including non-prompt areas, at later timesteps (Fig.~\ref{fig:problem_def}(b)).
High-ranked tokens in CFG importance signal, however, remain anchored to foreground objects throughout the entire timestep, confining its search for important tokens only to prompt-aligned regions.
As a result, CFG fails to identify the very tokens that should be protected: actively evolving tokens outside the foreground are prematurely merged into nearby or similar tokens, suppressing their distinct updates.

Second, CFG is insensitive to \textbf{prompt-independent structural importance}.
To produce natural and coherent images, diffusion models actively generate scene elements that are not directly mentioned in the prompt such as background objects and supporting structures~\cite{rombach2022high, peebles2023scalable}.
Although these prompt-independent structures contribute little to text guidance, they carry essential structural information such as boundaries between homogeneous regions, edges of background objects, and fine textures that anchor visual coherence.
% Although these prompt-independent structures contribute 
% little to text guidance, they \djkim{also} carry essential structural information.
However, because CFG only measures text-conditioned discrepancy, structural boundaries in such non-prompt regions tend to receive low scores across all timesteps (Fig.~\ref{fig:problem_def}(c)), making them vulnerable to merging into neighboring homogeneous areas.
Taken together, these observations indicate that token importance should not be defined solely by text relevance, but by diffusion dynamics: \emph{how unresolved a token remains} and \emph{how much structural information it carries} at the current denoising step~\cite{zhao2024dynamic}.

To faithfully capture diffusion dynamics, we define two 
complementary signals that quantify the spatiotemporal dynamics
of each token during denoising.
For temporal dynamics, we start from the score function~\cite{song2020score}, 
which quantifies how each token is updated at the current denoising step.
In particular, we project this score into clean-image space via the Tweedie 
denoised estimate~\cite{robbins1992empirical, efron2011tweedie}, 
and use its inter-step difference as the \emph{temporal signal}, assigning high values to tokens that are unresolved.
Unlike CFG, this score-derived signal reflects the actual update applied to every token, regardless of prompt relevance, providing an image-wide view of generative activity (Fig.~\ref{fig:problem_def}(b)).
For the \emph{structural signal}, we measure the residual 
between the Tweedie estimate and its coarse-scale smoothed version, which acts as a high-pass filter highlighting structural components throughout the image.
This captures prompt-independent structural boundaries, including weakly conditioned structures that CFG often underestimates.
By jointly modeling temporal evolution and structural saliency, we obtain a dynamics-aware token importance signal that guides merging decisions.

Building on this dynamics-aware importance, we propose \textbf{D}ynamics-\textbf{a}ware \textbf{T}oken \textbf{Me}rging \textbf{(DaTMe)}, a training-free token merging framework for diffusion transformers~\cite{peebles2023scalable,chen2023pixartalphafasttrainingdiffusion,flux2024}.
At each denoising step, DaTMe assigns token roles at the intersection of the temporal and structural signals.
High-scoring tokens are assigned as constraint tokens and excluded from merging as dynamically or structurally important regions, while low-scoring tokens are assigned as source tokens from relatively stable and homogeneous regions that can be safely merged.
To meet the target compression budget, DaTMe supplements the source set with grid-based spatial sampling, preventing local collapse.
The resulting source tokens are then merged only within nearby regions via locality-constrained matching, built upon~\cite{Wu_2025_ICCV}.
We primarily evaluate DaTMe on PixArt-$\alpha$ and FLUX, where it outperforms prior methods while maintaining visual fidelity across the entire image. 
Our contributions are as follows:
\begin{itemize}[itemsep=4pt, topsep=3pt, parsep=0pt, partopsep=0pt]
    \item We introduce a dynamics-aware importance map derived from the Tweedie estimate, jointly capturing the temporal evolution and structural saliency of each token.
    \item We propose \textbf{DaTMe}, a training-free token merging framework that leverages the proposed temporal and spatial importance signals to identify safely mergeable tokens while preserving structurally critical regions.
    \item We validate DaTMe on DiT (PixArt-$\alpha$) and MM-DiT (FLUX), showing strong preservation of image-wide details and state-of-the-art performance across most metrics.
\end{itemize}

\section{Related Works}

\noindent\textbf{Token Reduction in Large Models.}
Token reduction is a widely used strategy for improving Transformer efficiency, mainly through token pruning or token merging~\cite{bolya2023token,rao2021dynamicvit,chen2023diffrate,yang2025libra,zhang2025tf}.
% Pruning directly removes less informative tokens, whereas merging reduces sequence length by combining similar tokens and can better preserve information~\cite{bolya2022token}.
Representative methods such as ToMe~\cite{bolya2022token} use training-free bipartite matching, while later approaches refine token selection using attention or frequency-domain cues~\cite{yang2025libra,zhang2025tf}.
However, most of these methods target single-forward-pass architectures with relatively static token importance. 
Diffusion models, by contrast, update token representations over many steps, motivating diffusion-specific approaches such as ToMeSD~\cite{bolya2023token}.

\noindent\textbf{Token Merging in Diffusion Models.}
A wide range of methods have been proposed to accelerate diffusion models, including improved samplers~\cite{luo2023latent,salimans2022progressive,meng2023distillation,song2020denoising,song2023consistency} and weight quantization~\cite{li2023q,shang2023post}. However, such approaches often require additional training or optimization, introducing non-trivial overhead.
Training-free methods instead reduce the number of tokens processed during inference through pruning~\cite{zhao2024dynamic,kim2024token}, caching~\cite{ma2024deepcache,wimbauer2024cache,liu2024faster}, or merging~\cite{Wu_2025_ICCV,bolya2023token,zhang2025training,lu2025toma,fang2025attend}.
Among them, ToMeSD~\cite{bolya2023token} extends token merging in ViTs~\cite{bolya2022token} to diffusion models by grouping tokens based on feature similarity.
Several subsequent methods~\cite{lu2025toma,fang2025attend} leverage cross-attention maps to identify important tokens from semantically or structurally salient regions, aiming not merely to reduce computation but also to preserve informative representations during token reduction~\cite{kong2025token}.
More recently,~\cite{Wu_2025_ICCV} uses classifier-free guidance (CFG) as a saliency signal without additional attention computation, but because CFG primarily reflects alignment with the conditioning signal, it fails to capture the spatially heterogeneous and temporally evolving dynamics of the full image.
In contrast, our method leverages the Tweedie denoised estimate to construct a dynamics-aware importance map that captures both temporal evolution and spatial structure across the image.
\section{Method}
% \noindent\textbf{Overview.}
% \djkim{Our goal is to merge token for efficient image generation. In our method, DaTMe, we merge define tokens with low tweedie estimate difference and low frequency as important tokens. In Sec.3.1, blabla~~)}
Our goal is to design a token merging strategy for efficient image generation by accounting for diffusion dynamics.
To this end, we first define a \textbf{dynamics-aware importance signal} that characterizes token importance from two complementary perspectives: \textbf{temporal dynamics} and \textbf{structural saliency}.
Based on this importance map, we assign tokens into functional roles for merging, distinguishing safe merge sources and structurally important constraint tokens.
We then perform locality-constrained token merging, together with grid-based source completion to satisfy the target compression ratio.
In the following, we first motivate our Tweedie-based spatiotemporal importance signals (Sec.~\ref{sec:whytweedie}), and then describe how they are used in our proposed DaTMe framework (Sec.~\ref{sec:datme}).

\subsection{Preliminaries}

\noindent\textbf{Score Function.}
In diffusion models~\cite{rombach2022high, peebles2023scalable, podell2023sdxl}, the forward process gradually corrupts a clean
image $x_0$ by adding Gaussian noise over $t\in[0,T]$ timesteps:
\begin{equation}
    x_t = \alpha_t x_0 + \sigma_t \epsilon, \quad \epsilon \sim \mathcal{N}(0, I),
\end{equation}
% where $\alpha_t$ and $\sigma_t$ denote the signal and noise scales
% at timestep $t$, respectively.
where $\alpha_t$ and $\sigma_t$ denote the signal and noise scales at timestep $t$, respectively, satisfying $\alpha_t^2 + \sigma_t^2 = 1$.
The reverse process aims to recover $x_0$ from $x_T$ by iteratively
denoising, guided by the score function $\nabla_{x_t} \log p_t(x_t)$,
which represents the gradient of the log-density of the noisy
distribution $p_t(x_t)$ with respect to $x_t$~\cite{song2020score}.
Intuitively, the score function defines a vector field that points
toward regions of higher data likelihood, guiding the noisy sample
toward the clean data manifold throughout the denoising process.
In practice, the score function is estimated via a neural network
$\epsilon_\theta$ trained to predict the noise $\epsilon$ added at
each timestep. The score is then obtained from the predicted noise as:
\begin{equation}
    s(t) = \nabla_{x_t} \log p_t(x_t) \approx -\frac{1}{\sigma_t}
    \epsilon_\theta(x_t, t, c),
    \label{eq:score}
\end{equation}
where $c$ denotes the conditioning signal (e.g., text prompt).
% Importantly, this score estimate is already computed during standard 
% denoising and is therefore available at every inference step without 
% additional cost.

\noindent\textbf{Tweedie's Estimation.}
Tweedie's formula~\cite{efron2011tweedie} provides a closed-form
expression for the posterior mean of the clean sample $x_0$ given
the noisy observation $x_t$:
\begin{equation}
    \hat{x}_0(t) = \mathbb{E}[x_0 \mid x_t]
    = \frac{x_t + \sigma_t^2 \nabla_{x_t} \log p_t(x_t)}{\alpha_t}.
\end{equation}
Substituting the learned score estimator from Eq.~\ref{eq:score}, this becomes:
\begin{equation}
    \hat{x}_0(t) \approx \frac{x_t - \sigma_t\,\epsilon_\theta(x_t, t, c)}{\alpha_t},
    \label{eq:tweedie}
\end{equation}
% which corresponds to the denoised estimate of $x_0$ at timestep $t$,
% commonly referred to as the \textit{Tweedie's denoised estimate}.
which gives a denoised estimate of $x_0$ at timestep $t$, termed the \textit{Tweedie denoised estimate}.
Unlike the noisy sample $x_t$, which is directly modulated 
by the noise level $\sigma_t$, $\hat{x}_0(t)$ provides an 
estimate in the predicted clean image space~\cite{rout2024beyond}, 
offering a noise-normalized view of the current 
generation state.
% Unlike the noisy sample $x_t$, which operates in noise space and is
% directly modulated by $\sigma_t$, $\hat{x}_0(t)$ lives in the 
% predicted clean image space~\cite{rout2024beyond} and provides a 
% noise-scale-decoupled view of the current generation state.

\subsection{Spatiotemporal Tweedie Importance Signal}
\label{sec:whytweedie}
\begin{figure}[t]
    \centering
    \includegraphics[width=1.0\linewidth]{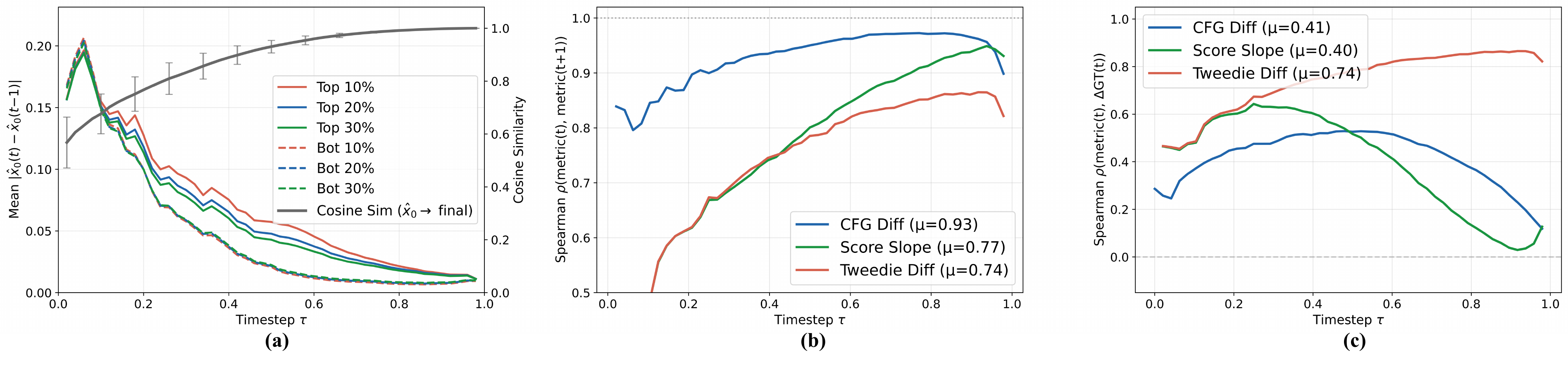}
\caption{
\textbf{(a) Cosine similarity and temporal trajectories of top- and bottom-ranked tokens.} 
The Tweedie estimate progressively approaches the final latent over denoising. Meanwhile, tokens with high pseudo-GT magnitude $|x_0-\hat{x}_0(t)|$ exhibit consistently larger changes than low-ranked tokens. This indicates that the pseudo-GT effectively separates active tokens from converged ones.
% The Tweedie estimate progressively approaches the final latent over denoising, while tokens with high pseudo-GT magnitude $|x_0-\hat{x}_0(t)|$ exhibit consistently larger changes than low-ranked tokens, indicating that the pseudo-GT effectively separates active tokens from converged ones.
\textbf{(b) Step-wise autocorrelation of token-importance signals.} Computed between consecutive timesteps, lower values indicate more dynamic importance.
The proposed Tweedie-based signal exhibits lower autocorrelation than CFG and score slope.
(\textbf{c) Correlation with the pseudo-GT change magnitude.} The proposed Tweedie difference remains better aligned with actual token changes across timesteps than CFG or score slope. $\tau$ denotes the normalized timestep, scaled to the range $[0,1]$.
}
    \label{fig:autocorr}
\end{figure}

\noindent\textbf{Theoretical Comparison of Temporal Signals.}
% To capture token-level evolution during the iterative denoising process, we focus on the score function~\cite{song2020score}, which reflects image-wide changes throughout generation.
% We first compare two temporal signals derived from it: the naive score slope, $|s(t)-s(t-1)|$, and the Tweedie difference, $|\hat{x}_0(t)-\hat{x}_0(t-1)|$, by analyzing what information each signal conveys across denoising steps.
To characterize token-level evolution during denoising, we compare two temporal signals derived from the score function~\cite{song2020score}: the raw score slope, $|s(t)-s(t-1)|$, and the Tweedie difference, $|\hat{x}_0(t)-\hat{x}_0(t-1)|$.
As derived in Appendix~\ref{sec:formula}, the score slope can be decomposed into two components:
\begin{equation}
s(t) - s({t-1}) 
= \underbrace{\frac{\epsilon_\theta(t-1) - \epsilon_\theta(t)}{\sigma_{t-1}}}_{\text{difference term}} 
+ \underbrace{\epsilon_\theta(t) \left( \frac{1}{\sigma_{t-1}} - \frac{1}{\sigma_t} \right)}_{\text{scale term}}.
\end{equation}
The first term reflects changes in the model prediction across timesteps and is therefore associated with actual token-level evolution.
% In contrast, the second term arises purely from the predefined global noise schedule $\sigma_t$ and does not directly encode meaningful structural change.
In contrast, the second term arises purely from the predefined global noise schedule $\sigma_t$ rather than changes in the model prediction itself.
As denoising approaches the late steps ($\sigma_t \to 0$), the inverse difference $\left| \frac{1}{\sigma_{t-1}} - \frac{1}{\sigma_t} \right|$ grows rapidly.
This suggests that the scale term can increasingly dominate the score slope in later stages, making the resulting signal less reliable.
% as a measure of token evolution.
% As denoising approaches the late steps, $\sigma_t$ becomes small, causing the schedule-driven term to grow rapidly and eventually dominate the score slope.
To obtain a cleaner temporal measure, we instead project the score into predicted clean-image space via the Tweedie estimate $\hat{x}_0(t)$~\cite{rout2023solving} and consider its inter-step residual, $\hat{x}_0(t)-\hat{x}_0(t-1)$.
By algebraic decomposition (Appendix~\ref{sec:formula}), this residual reduces to:
\begin{equation}
    \hat{x}_0(t) - \hat{x}_0(t-1) 
    = \frac{\sigma_{t-1}}{\alpha_{t-1}}\big(\epsilon_\theta(t-1) - \epsilon_\theta(t)\big),
\end{equation}
which removes the scale term that appears in the raw score slope and retains only the difference term update.
Furthermore, the update is scaled by $\frac{\sigma_{t-1}}{\alpha_{t-1}}$, and this factor decreases as denoising concludes, naturally attenuating the signal in late steps.
These properties make the Tweedie difference a more stable temporal signal for measuring token evolution.

\noindent\textbf{Empirical Validation of Temporal Signals.}
We validate the theoretical analysis above through two empirical tests: temporal autocorrelation and correlation with a pseudo-GT change measure.
\begin{wrapfigure}[24]{r}{0.48\linewidth}
\vspace{-8pt}
\centering
\includegraphics[width=\linewidth]{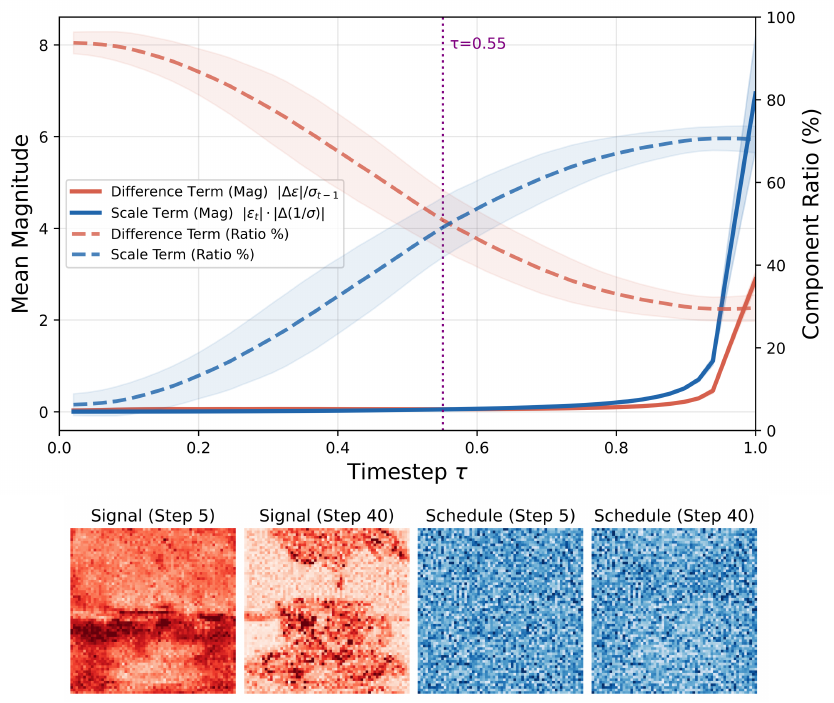}
% \vspace{-8pt}
\caption{
\textbf{Difference- and Scale- components of the score slope.}
(Top) Mean magnitude and contribution ratio of the two terms: the scale term becomes dominant in late steps. (Bottom) Spatial distributions: the difference term captures structured token-level changes, whereas the scale term remains spatially unstructured.
}
\label{fig:decomp}
\vspace{-8pt}
\end{wrapfigure}
We first examine temporal autocorrelation to assess whether each signal remains responsive to token-wise refinement over denoising steps (Fig.~\ref{fig:autocorr}(b)).
The CFG-based importance~\cite{Wu_2025_ICCV} exhibits consistently high autocorrelation across all timesteps, confirming its static nature and its bias toward fixed, prompt-aligned foreground regions, as previously observed in Fig.~\ref{fig:problem_def}(b).
In contrast, both score-based signals, namely the raw score slope and the Tweedie difference, initially show noticeably lower autocorrelation, indicating greater sensitivity to image-wide refinement than CFG.
However, temporal autocorrelation alone does not naturally imply that a signal reflects token-level convergence.

To evaluate whether a signal faithfully reflects token-level convergence, we measure its correlation with a reference signal constructed from the per-timestep Tweedie estimate $\hat{x}_0(t)$, which we treat as a proxy for the clean image following~\cite{rout2023solving}.
We then define the deviation between the final clean latent $x_0$ and $\hat{x}_0(t)$, denoted as $|x_0-\hat{x}_0(t)|$, as a pseudo-GT measure of how much each token is still expected to change.
As shown in Fig.~\ref{fig:autocorr}(a), the cosine similarity between $\hat{x}_0(t)$ and the final denoised latent steadily approaches $1.0$ over time, indicating that $\hat{x}_0(t)$ progressively approximates the final generation state.
Moreover, when tokens are partitioned by their pseudo-GT magnitude, high-ranked tokens consistently exhibit larger changes across denoising steps than low-ranked ones, suggesting that this pseudo-GT provides a meaningful per-token reference for remaining change.

Using this pseudo-GT, a clear difference between the two score-based signals emerges (Fig.~\ref{fig:autocorr}(c)).
While both signals show high correlation with the pseudo-GT in the early stages, the correlation of the raw score slope drops sharply around normalized timestep $0.4$--$0.6$.
Consistent with our theoretical decomposition, this degradation closely matches the range where the scale term empirically overtakes the difference term (Fig.~\ref{fig:decomp}).
At the same time, the autocorrelation of the score slope rises (Fig.~\ref{fig:autocorr}(b)), indicating increasingly static behavior and reduced sensitivity to token-wise refinement.
In contrast, by removing scale term interference, the Tweedie difference maintains both a higher correlation with the pseudo-GT and a consistently lower temporal autocorrelation throughout denoising.
Taken together, these results suggest that the Tweedie difference provides a more reliable temporal signal for tracking token evolution than either the raw score slope or CFG.

\noindent\textbf{Tweedie Importance Map.}
Building on this analysis, we define the temporal importance map as the inter-step residual in predicted clean-image space:
\begin{equation}
    \mathcal{M}_{temp}^{(t)} = \left|\hat{x}_0(t) - \hat{x}_0(t-1)\right|,
\end{equation}
which quantifies how much the predicted clean structure is still evolving at each spatial location.
This temporal map captures when each token is actively changing during denoising.

However, as previously observed in Fig.~\ref{fig:problem_def}(c), 
CFG also fails to protect structurally salient tokens in 
non-prompt regions—those carrying essential boundary and texture 
information. 
Under CFG-based importance, such tokens consistently 
receive low scores throughout denoising, leading them to be 
mistakenly treated as safe merge candidates and causing object 
boundaries to be diluted (Fig.~\ref{fig:signal_effect}).
% This indicates that token importance depends not only on temporal 
% evolution, but also on how structurally distinct a token is in 
% the clean image estimate.

To capture this structural property, we introduce a spatial axis derived from the same Tweedie estimate.
% Specifically, we focus on distinguishing structurally salient tokens that appear temporally stable from genuinely mergeable regions, so that only locally homogeneous, low-frequency regions are selected as safe merge targets.
Specifically, we use the spatial axis to distinguish structurally salient but temporally stable tokens from genuinely mergeable homogeneous regions.
% To this end, we construct a coarse-scale approximation of $\hat{x}_0(t)$ through interpolation-based downsampling followed by upsampling~\cite{koh2025scalediff}, and define the spatial axis as the residual between this smoothed representation and the original estimate:
To this end, we define the spatial axis as the residual between $\hat{x}_0(t)$ and its coarse-scale approximation obtained via interpolation-based transformation~\cite{koh2025scalediff}:
\begin{equation}
    \mathcal{M}_{spat}^{(t)} = \left|\hat{x}_0(t) - \mathrm{UpSample}_{\times s}\!\Big(\mathrm{DownSample}_{\times s}\!\big(\hat{x}_0(t)\big)\Big) \right|,
    \label{eq:spatial}
\end{equation}
where $s$ denotes the spatial scale used to construct the coarse approximation.
% This operation acts as a simple low-pass decomposition, where the residual highlights high-frequency structures.
This spatial map assigns high responses to contours, boundaries, and fine textures, while giving low responses to locally homogeneous regions.
% This operation suppresses coarse homogeneous content and highlights structurally distinct regions in the residual.
% Accordingly, $\mathcal{M}_{spat}^{(t)}$ emphasizes contours, boundaries, and fine textures while assigning low responses to homogeneous background regions.
% Accordingly, $\mathcal{M}_{spat}^{(t)}$ emphasizes contours, boundaries, and fine textures while assigning low responses to locally homogeneous regions.
By combining the temporal and spatial axes, we obtain an importance signal that reflects both how tokens evolve over time and how much structural information they carry at each denoising step.

\subsection{Dynamics-aware Token Merging (DaTMe)}
\label{sec:datme}
At each timestep $t$, given a target compression ratio $r$, we denote the full token set as $\mathcal{S}_{tot}$ with $N = |\mathcal{S}_{tot}|$, and aim to merge $n_{tgt}=\lfloor rN \rfloor$ tokens.
% \djkim{(how about full token set $S_{tot}$?)}
While feature-similarity-based merging strategies work well in U-Net-based architectures~\cite{rombach2022high,podell2023sdxl}, directly applying them to Diffusion Transformers (DiTs)~\cite{peebles2023scalable,esser2024scaling} often causes severe structural distortion~\cite{Wu_2025_ICCV,lu2025toma}.
To prevent such distortion, we propose Dynamics-aware Token Merging (DaTMe), which performs three steps:
(1) assigning token roles based on the proposed 
\begin{wrapfigure}[16]{r}{0.48\linewidth}
\vspace{-8pt}
\centering
\includegraphics[width=\linewidth]{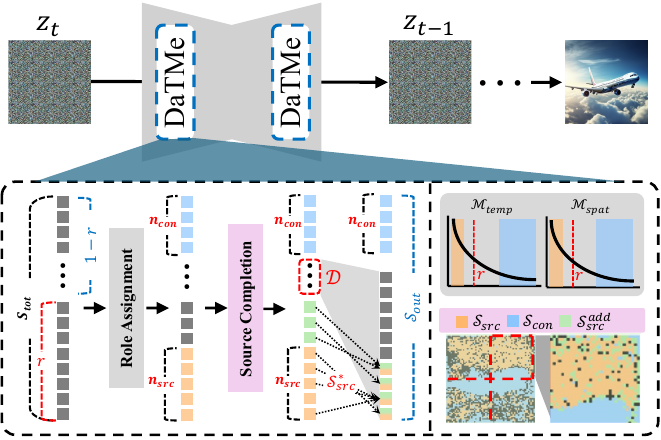}
% \vspace{-8pt}
\caption{\textbf{Proposed DaTMe pipeline.} Using the spatiotemporal axes, DaTMe assigns token roles and performs merging within diffusion layers.}
\label{fig:pipe}
\vspace{-10pt}
\end{wrapfigure}
spatiotemporal importance maps,
(2) completing the source budget in a spatially distributed manner, and (3) performing locality-constrained spatial matching.
A detailed step-by-step algorithm of DaTMe is provided in the Appendix~\ref{sec:datme_detail}.

\noindent\textbf{Dynamics-Aware Role Assignment.} 
Rather than imposing a hand-crafted cutoff for what should be 
considered important, DaTMe adaptively assigns token roles at 
each denoising step based on their temporal and spatial responses 
under the given compression ratio $r$.
% We hypothesize that source tokens should come from regions that 
% are both temporally stable and spatially homogeneous, since 
% merging temporally evolving tokens disrupts ongoing refinement 
% while merging structurally salient tokens blurs local detail.
To prevent quality degradation caused by merging temporally 
evolving or structurally salient tokens, we first define a 
conservative source set $\mathcal{S}_{src}\subset  \mathcal{S}_{tot}$ as the 
intersection of the lowest-$r$ fractions of tokens ranked by 
$\mathcal{M}_{temp}$ and $\mathcal{M}_{spat}$.
Conversely, we define a constraint set, $\mathcal{S}_{con}\subset  \mathcal{S}_{tot}$, as the intersection of the top-$(1-r)$ fractions of tokens on both axes, preserving tokens that are both temporally unstable and structurally salient.
By preserving $\mathcal{S}_{con}$ throughout merging, we use them as structural anchors that keep critical boundaries from dissolving into surrounding regions.

\noindent\textbf{Spatially Distributed Source Completion.}
Because $\mathcal{S}_{src}$ is defined by the strict intersection of two conditions, its size $n_{src} = |\mathcal{S}_{src}|$ is typically smaller than the target budget $n_{tgt}$ (i.e., $n_{src} < n_{tgt}$).
We therefore select an additional set $\mathcal{S}_{src}^{add}$ from the remaining unassigned tokens to satisfy $|\mathcal{S}_{src} \cup \mathcal{S}_{src}^{add}| = n_{tgt}$.
To avoid concentrating these additional source tokens in a specific region—which would lead to local information loss after merging—we adopt a grid-based spatial sampling strategy.
Specifically, we reshape the token sequence into a spatial 
feature map, partition it into $P$ non-overlapping grid cells, 
and uniformly sample $(n_{tgt} - n_{src})/P$ tokens from each cell.
The final source set is then given by $\mathcal{S}_{src}^* = \mathcal{S}_{src} \cup \mathcal{S}_{src}^{add}$.

\noindent\textbf{Locality-Constrained Routing.} 
% The remaining tokens---those not assigned to 
% $\mathcal{S}_{src}^*$ or $\mathcal{S}_{con}$---form 
% the destination pool $\mathcal{D}\subset  \mathcal{S}_{tot}$, into which 
% $\mathcal{S}_{src}^*$ is merged via bipartite soft matching.
The remaining tokens, those not assigned to $\mathcal{S}_{src}^*$ or $\mathcal{S}_{con}$, form the destination pool $\mathcal{D}\subset  \mathcal{S}_{tot}$, into which $\mathcal{S}_{src}^*$ is merged via bipartite soft matching.  
Because DiTs process tokens as a sequence without inherent spatial inductive bias~\cite{d2021convit}, relying purely on feature similarity can collapse spatially distant tokens and distort local spatial structure.
We therefore use spatial adjacency as the matching 
criterion~\cite{Wu_2025_ICCV}, pairing each source token with its 
spatially nearest destination and obtaining $\mathcal{S}_{merged}=\mathrm{Merge}(\mathcal{D},\mathcal{S}_{src}^*)$.
The final output token set is $\mathcal{S}_{out}=\mathcal{S}_{merged}\cup\mathcal{S}_{con}$, with 
$|\mathcal{S}_{out}|=N-n_{tgt}$.
% |\mathcal{S}_{out}|=(1-r)\cdot N$.
By excluding $\mathcal{S}_{con}$ from both pools and 
restricting merging to local neighborhoods, DaTMe compensates 
for the weak spatial inductive bias of DiTs and preserves 
visual fidelity throughout merging.

\section{Experiments}

%TODO
\begin{figure*}[t]
\centering

\begin{minipage}[t]{0.49\linewidth}
    \vspace{0pt}
    \centering
    \includegraphics[width=\linewidth]{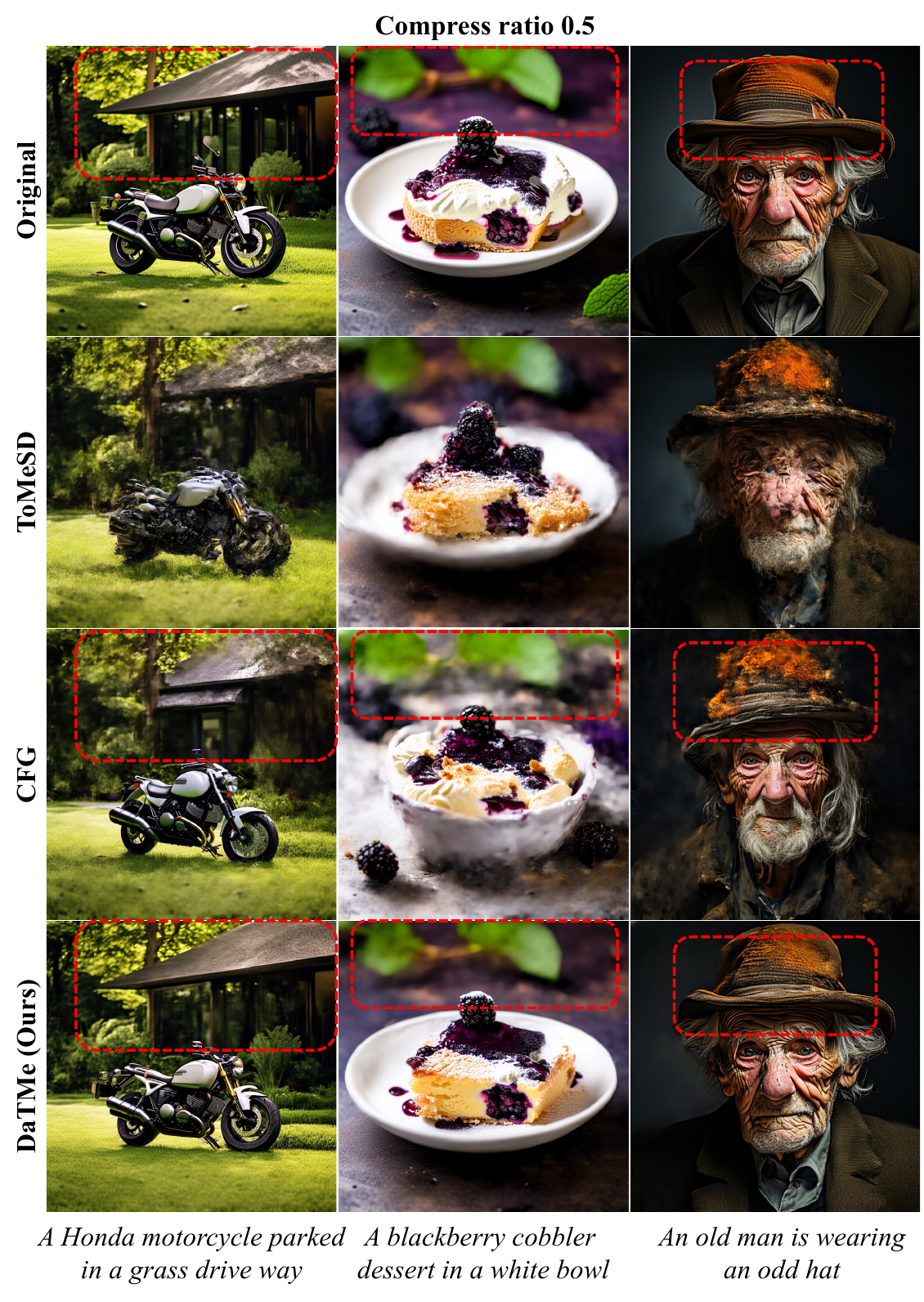}
    \captionof{figure}{\textbf{Qualitative comparison under compression ratio 0.5 on PixArt-$\alpha$.}
    DaTMe better preserves structural details and background content.}
    \label{fig:pixart_compare}
\end{minipage}
\hfill
\begin{minipage}[t]{0.49\linewidth}
    \vspace{0pt}
    \centering
    \includegraphics[width=\linewidth]{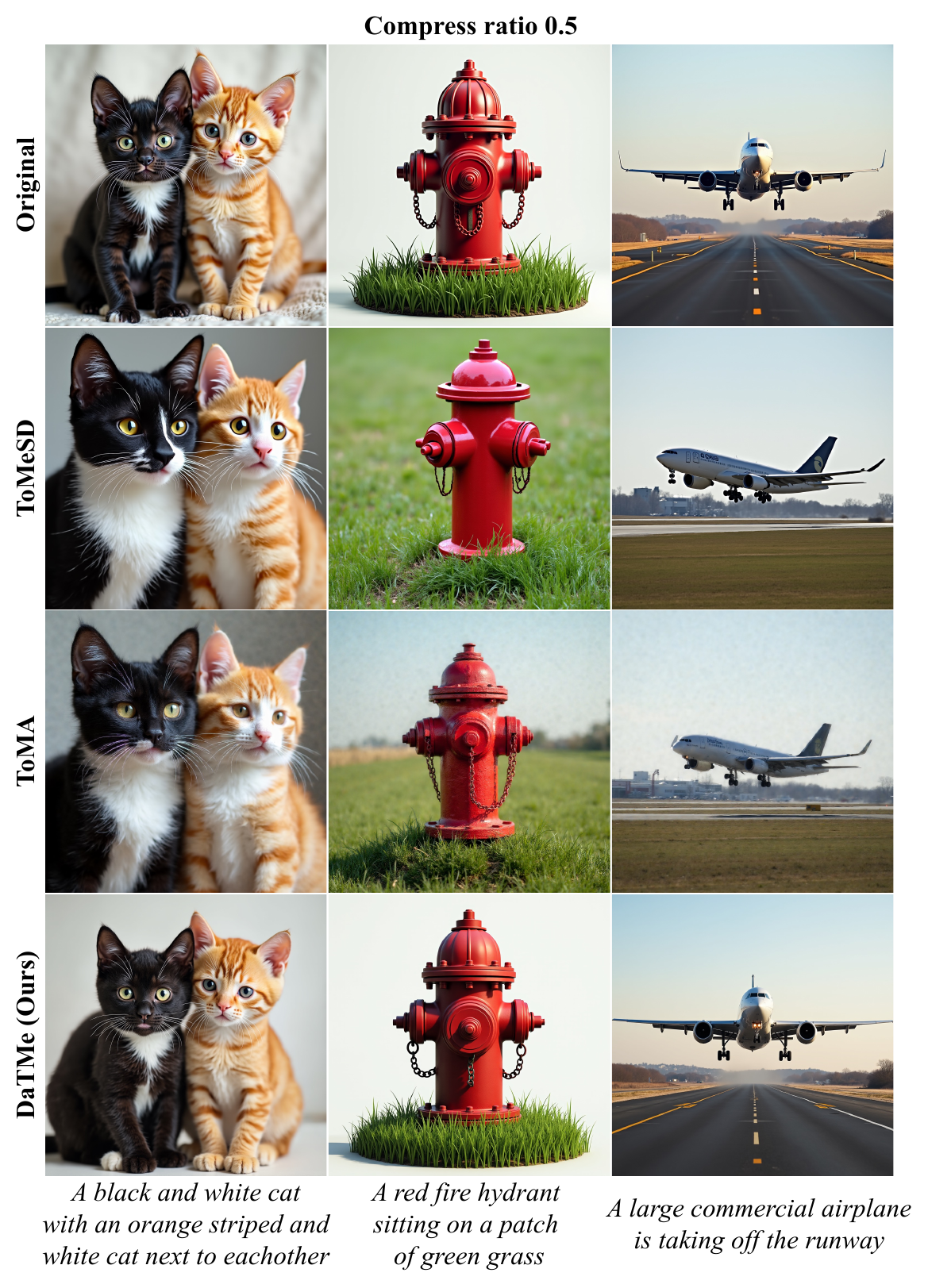}
    \captionof{figure}{\textbf{Qualitative comparison under compression ratio 0.5 on FLUX.}
    DaTMe better preserves structural details with lower quality degradation.}
    \label{fig:flux_compare}
\end{minipage}

\vspace{-8pt}
\end{figure*}

\noindent\textbf{Implementation Details and Evaluation Metrics.}
We primarily conduct experiments on PixArt-$\alpha$~\cite{chen2023pixartalphafasttrainingdiffusion} using 10k image-prompt pairs sampled from the COCO 2014 validation set~\cite{lin2014microsoft}.
We set the interpolation scale to $s=4$, and for grid-based source completion, we use $P=4$ grid cells.
We evaluate three compression ratios: 0.3, 0.5, and 0.7.
Following~\cite{Wu_2025_ICCV}, we apply DaTMe to the middle-range self-attention layers of the DiT~\cite{peebles2023scalable} backbone.
To further evaluate scalability on a larger DiT model, we apply DaTMe to FLUX~\cite{flux2024}.
We use FLUX.1-dev and additionally apply DaTMe from the middle block to the final block within the joint-attention blocks using the same hyperparameters.
More details are in the supplementary materials.
For evaluation, we follow prior work~\cite{Wu_2025_ICCV,bolya2023token,lu2025toma,fang2025attend} and report CLIP-T and FID.
In addition, we report HPSv2~\cite{wu2023human}, which complements CLIP-T by measuring preference-oriented prompt faithfulness and perceptual quality.
We also report SSIM, PSNR, and LPIPS to measure image-level fidelity and detail preservation relative to the original output.
We further report the Aesthetic score using Aesthetic Predictor V2.5, a SigLIP-based reference-free image aesthetic predictor.
All experiments are conducted on a single RTX PRO 6000 Blackwell.

\begin{table}[t]
\centering

\begin{minipage}[t]{0.55\linewidth}
\centering
\caption{\textbf{Text alignment and efficiency scores on PixArt-$\alpha$.}
DaTMe achieves the best overall quality-efficiency trade-off, preserving strong text alignment with minimal degradation under token compression.}
\label{tab:pixart_main_metrics}
\vspace{2pt}

\small
\setlength{\tabcolsep}{3.5pt}
\renewcommand{\arraystretch}{1.05}
\resizebox{\linewidth}{!}{
\begin{tabular}{l c c c c c}
\toprule
& \textbf{CLIP-T}$\uparrow$ & \textbf{HPSv2}$\uparrow$ & \textbf{FID}$\downarrow$ & \textbf{Latency}(s)$\downarrow$ & \textbf{TFLOPs}$\downarrow$ \\
\midrule
PixArt-$\alpha$ (Base)~\cite{chen2023pixartalphafasttrainingdiffusion} & 31.2485 & 29.2780 & 42.1001 & 1.9417 & 183.45 \\
\midrule
\multicolumn{6}{c}{\textbf{Compression ratio 0.3}} \\
\midrule
ToMeSD~\cite{bolya2023token} & 30.5735 & 24.1699 & 61.5229 & 1.6917 & 172.514 \\
CFG~\cite{Wu_2025_ICCV} & \underline{30.8801} & \underline{27.7375} & \underline{44.6070} & 1.7458 & 172.514 \\
DaTMe (Ours) & \textbf{31.1375} & \textbf{28.5591} & \textbf{42.4474} & 1.8030 & 175.631 \\
\midrule
\multicolumn{6}{c}{\textbf{Compression ratio 0.5}} \\
\midrule
ToMeSD & 29.5178 & 21.3334 & 94.8901 & 1.5918 & 165.201 \\
CFG & \underline{30.4645} & \underline{25.8345} & \underline{52.2912} &  1.6445 & 165.201 \\
DaTMe (Ours) & \textbf{31.0142} & \textbf{27.5174} & \textbf{43.4155} & 1.6884 & 170.410 \\
\midrule
\multicolumn{6}{c}{\textbf{Compression ratio 0.7}} \\
\midrule
ToMeSD & 28.2805 & 18.9797 & 123.0139 & 1.5356 & 157.897 \\
CFG & \underline{29.4543} & \underline{22.6156} & \underline{70.7062} & 1.5761 & 157.897 \\
DaTMe (Ours) & \textbf{30.6259} & \textbf{25.2393} & \textbf{47.0519} & 1.6426 & 165.193 \\
\bottomrule
\end{tabular}
}
\end{minipage}
\hfill
\begin{minipage}[t]{0.42\linewidth}
\centering
\caption{\textbf{Image fidelity score on PixArt-$\alpha$.} DaTMe achieves stronger structural preservation and image-level fidelity under token compression.}
\label{tab:pixart_fidelity_metrics}
\vspace{2pt}

\small
\setlength{\tabcolsep}{3.8pt}
\renewcommand{\arraystretch}{1.05}
\resizebox{\linewidth}{!}{
\begin{tabular}{l c c c c}
\toprule
& \textbf{SSIM}$\uparrow$ & \textbf{PSNR}$\uparrow$ & \textbf{LPIPS}$\downarrow$ & \textbf{Aes.}$\uparrow$ \\
\midrule
\multicolumn{4}{c}{\textbf{Compression ratio 0.3}} \\
\midrule
ToMeSD~\cite{bolya2023token} & \underline{0.7306} & \underline{18.9748} & 0.2433  &5.9906\\
CFG~\cite{Wu_2025_ICCV} & 0.7137 & 17.8157 & \underline{0.2369} &\underline{6.1521}\\
DaTMe (Ours) & \textbf{0.8112} & \textbf{20.5136} & \textbf{0.1501} &\textbf{6.2014}\\
\midrule
\multicolumn{4}{c}{\textbf{Compression ratio 0.5}} \\
\midrule
ToMeSD & \underline{0.6737} & \underline{17.7161} & \underline{0.3151} & 5.7739\\
CFG & 0.6331 & 16.2768 & 0.3254 & \underline{6.1135}\\
DaTMe (Ours) & \textbf{0.7300} & \textbf{18.4277} & \textbf{0.2258} & \textbf{6.1695}\\
\midrule
\multicolumn{4}{c}{\textbf{Compression ratio 0.7}} \\
\midrule
ToMeSD & \underline{0.6301} & \underline{16.8537} & \underline{0.3682} & 5.5809\\
CFG & 0.5695 & 15.2003 & 0.4129 & \underline{5.9559}\\
DaTMe (Ours) & \textbf{0.6581} & \textbf{16.9048} & \textbf{0.3078} & \textbf{6.0544}\\
\bottomrule
\end{tabular}
}
\end{minipage}

\end{table}

\begin{table}[!t]
\centering

\begin{minipage}[t]{0.55\linewidth}
\centering
\caption{\textbf{Text alignment and efficiency score on FLUX.}
DaTMe achieves stronger text alignment with comparable efficiency under token compression.
$\dagger$ denotes our reproduced results.}
\label{tab:flux_main_metrics}
\vspace{2pt}

\small
\setlength{\tabcolsep}{3.5pt}
\renewcommand{\arraystretch}{1.05}
\resizebox{\linewidth}{!}{
\begin{tabular}{l c c c c c}
\toprule
& \textbf{CLIP-T}$\uparrow$
& \textbf{HPSv2}$\uparrow$
& \textbf{FID}$\downarrow$
& \textbf{Latency}(s)$\downarrow$
& \textbf{TFLOPs}$\downarrow$ \\
\midrule
FLUX.1-dev (Base)~\cite{flux2024}
& 31.2274 & 29.6927 & 30.6141 & 14.601 & 2098 \\

\midrule
\multicolumn{6}{c}{\textbf{Compression ratio 0.3}} \\
\midrule
ToMeSD$^{\dagger}$~\cite{bolya2023token}
&\underline{31.3254}& \underline{28.9339}& \textbf{30.4576}&13.866 & 2046 \\

ToMA~\cite{lu2025toma}
& 31.1787 & 28.5153 & \underline{30.6520} & 13.790 & 1971 \\

DaTMe (Ours)
& \textbf{31.3524} & \textbf{29.5042} & 30.9258 & 13.772 & 2033 \\

\midrule
\multicolumn{6}{c}{\textbf{Compression ratio 0.5}} \\
\midrule
ToMeSD$^{\dagger}$
&31.1852& \underline{28.3595}& \textbf{30.8491}&13.278 & 2000  \\

ToMA
& \underline{31.2036} & 27.7566 & \underline{30.9883} & 13.288 & 1895 \\

DaTMe (Ours)
& \textbf{31.2621} & \textbf{28.9266} & 31.3423 & 13.271 & 1990 \\

\midrule
\multicolumn{6}{c}{\textbf{Compression ratio 0.7}} \\
\midrule
ToMeSD$^{\dagger}$
&31.2045& \underline{27.6455}& \textbf{31.0033}&12.728 &1954  \\

ToMA
&\underline{31.2191}& 26.4796&35.0541 & 12.775& 1793 \\

DaTMe (Ours)
&\textbf{31.2415}&\textbf{27.9093}& \underline{31.1585}&12.737  &1946  \\

\bottomrule
\end{tabular}
}
\end{minipage}
\hfill
\begin{minipage}[t]{0.42\linewidth}
\centering
\caption{\textbf{Image fidelity score on FLUX.}
DaTMe achieves stronger structural preservation and image fidelity.}
\label{tab:flux_fidelity_metrics}
\vspace{2pt}

\small
\setlength{\tabcolsep}{3.8pt}
\renewcommand{\arraystretch}{1.05}
\resizebox{\linewidth}{!}{
\begin{tabular}{l c c c c}
\toprule
& \textbf{SSIM}$\uparrow$
& \textbf{PSNR}$\uparrow$
& \textbf{LPIPS}$\downarrow$
& \textbf{Aes.}$\uparrow$\\
\midrule

\multicolumn{4}{c}{\textbf{Compression ratio 0.3}} \\
\midrule
ToMeSD$^{\dagger}$~\cite{bolya2023token}
&0.6402&17.5070 &0.2616 & 5.7626\\

ToMA~\cite{lu2025toma}
& \underline{0.7415} & \underline{20.6450} & \underline{0.1844} & \underline{6.0210} \\

DaTMe (Ours)
& \textbf{0.7737} & \textbf{21.6352} & \textbf{0.1694} & \textbf{6.0318} \\

\midrule
\multicolumn{4}{c}{\textbf{Compression ratio 0.5}} \\
\midrule
ToMeSD$^{\dagger}$
&0.6166&16.9059 &0.2932 &5.6665 \\

ToMA
& \underline{0.7009} & \underline{20.0375} & \underline{0.2305} & \underline{5.9030}\\

DaTMe (Ours)
& \textbf{0.7087} & \textbf{20.2021} & \textbf{0.2176} & \textbf{6.0191}\\

\midrule
\multicolumn{4}{c}{\textbf{Compression ratio 0.7}} \\
\midrule
ToMeSD$^{\dagger}$
&0.6032 &16.6127& 0.3181 &5.5739 \\

ToMA
&\underline{0.6280}& \underline{17.6972}& \textbf{0.2835} &\underline{5.6971} \\

DaTMe (Ours)
&\textbf{0.6378}& \textbf{17.7655} & \underline{0.2877} &\textbf{5.9147} \\

\bottomrule
\end{tabular}
}
\end{minipage}

\end{table}

\subsection{Main Experiments}
\noindent\textbf{Qualitative Results.}
In Fig.~\ref{fig:pixart_compare} and Fig.~\ref{fig:flux_compare}, we compare qualitative results at a compression ratio of 0.5.
On PixArt-$\alpha$, directly merging tokens based on feature similarity (ToMeSD) severely disrupts spatial structure, often leading to obvious structural distortion.
Although the CFG-based method preserves foreground structures that are explicitly described in the prompt, it shows noticeably weaker preservation in background and other weakly conditioned regions, as highlighted by the red boxes.
This qualitatively supports our analysis that CFG-based importance is strongly biased toward prompt-conditioned content and is therefore less effective at preserving the full image context.

On FLUX, existing methods generally preserve the coarse prompt semantics, but still introduce noticeable deviations from the original generation.
ToMeSD tends to alter the original composition, object geometry, or viewpoint, as seen in the relative arrangement of the cats, the fire hydrant structure, and the airplane viewpoint.
ToMA shows similar structural drift while additionally introducing noisy visual degradation that weakens overall image quality.
In contrast, DaTMe more faithfully preserves the original spatial configuration and object structure while maintaining the intended prompt semantics.
These results demonstrate that DaTMe reduces both structural deviation and visual degradation across different architectures.
Additional results under various compression ratios are in Appendix~\ref{sec:add_exp}.

\noindent\textbf{Quantitative Results.} 
In Tab.~\ref{tab:pixart_main_metrics} and Tab.~\ref{tab:pixart_fidelity_metrics}, we compare DaTMe with prior methods on PixArt-$\alpha$~\cite{chen2023pixartalphafasttrainingdiffusion} under various compression ratios.
Although DaTMe is slightly slower than prior methods in terms of latency, the prior methods suffer from substantial quality degradation as the compression ratio increases, particularly in FID.
In contrast, DaTMe remains much more robust under compression, consistently preserving image quality and achieving the most favorable overall quality-efficiency trade-off.
Notably, DaTMe also attains substantially stronger structural preservation, which indicates that the proposed spatiotemporal importance signal effectively prevents structurally important tokens from being merged prematurely (Tab.~\ref{tab:pixart_fidelity_metrics}).

In the FLUX results, CFG-based token selection is not applicable because FLUX does not employ CFG, whereas DaTMe remains directly applicable despite the architectural difference.
Across all compression ratios, DaTMe consistently achieves the strongest CLIP-T and HPSv2 scores while maintaining comparable inference efficiency (Tab.~\ref{tab:flux_main_metrics}).
Although ToMeSD preserves reasonable text--image alignment, its substantially lower SSIM and PSNR and higher LPIPS indicate larger structural deviations from the original generation.
ToMA better preserves the original structure than ToMeSD, but still exhibits noticeable visual degradation, particularly in fine details and overall image quality.
In contrast, DaTMe achieves the strongest overall image fidelity, consistently improving SSIM, PSNR, and aesthetic quality while maintaining strong prompt alignment (Tab.~\ref{tab:flux_fidelity_metrics}).
These quantitative trends are consistent with Fig.~\ref{fig:flux_compare}, where DaTMe more faithfully preserves the original image structure.

\subsection{Additional Analysis}
\label{sec:ablation}

\begin{figure}[t]
\centering

% ===================== LEFT COLUMN =====================
\begin{minipage}[t]{0.42\linewidth}
    \vspace{0pt}
    \centering

    % ---------- Top: table ----------
    \captionof{table}{\textbf{Ablation on temporal and spatial axes.}
    We compare using only the temporal axis, only the spatial axis, both axes, and both axes with score slope instead of Tweedie difference.}
    \label{tab:axis_ablation}
    \vspace{2pt}

    \footnotesize
    \setlength{\tabcolsep}{3.8pt}
    \renewcommand{\arraystretch}{1.02}
    \resizebox{\linewidth}{!}{
    \begin{tabular}{l c c c c}
    \toprule
    \textbf{Setting} & \textbf{CLIP-T}$\uparrow$ & \textbf{SSIM}$\uparrow$ & \textbf{PSNR}$\uparrow$ & \textbf{FID}$\downarrow$ \\
    \midrule
    CFG~\cite{Wu_2025_ICCV} & 30.8801 & 0.7137 & 17.8157 & 44.6070 \\ \hline
    Only $\mathcal{M}_{\mathrm{temp}}$ & 31.0355 & 0.8109 & 20.5021 & \underline{42.5269} \\
    Only $\mathcal{M}_{\mathrm{spat}}$ & \underline{31.1318} & \underline{0.8110} & \underline{20.5029} & 42.5736 \\
    Ours (score slope) & 31.1166 & 0.8028 & 20.3944 & 42.9554 \\
    Ours & \textbf{31.1375} & \textbf{0.8112} & \textbf{20.5136} & \textbf{42.4474} \\
    \bottomrule
    \end{tabular}
    }

    \vspace{8pt}

    % ---------- Bottom: extra figure ----------
    \includegraphics[width=0.85\linewidth]{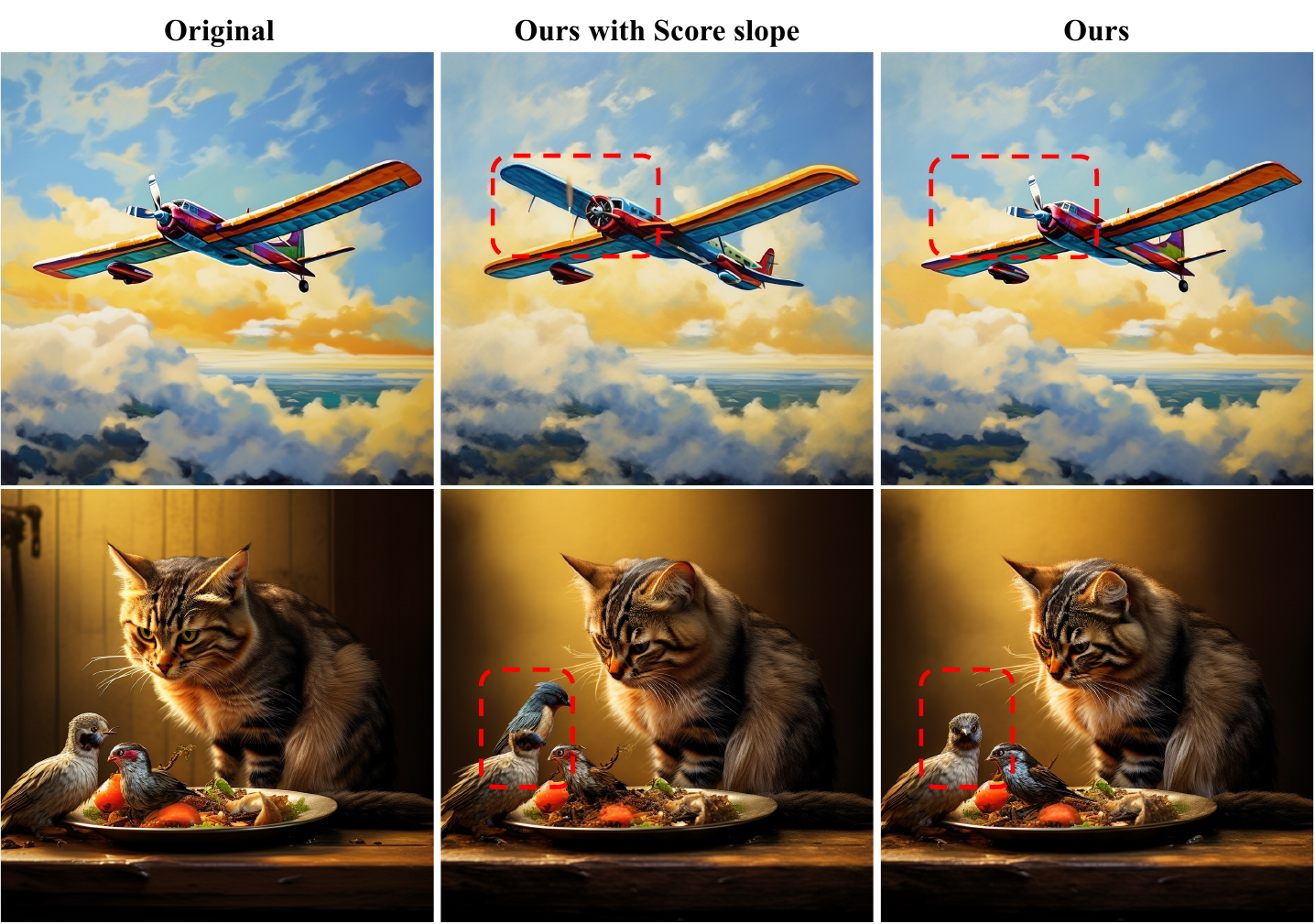}
    \vspace{2pt}
    \captionof{figure}{\textbf{Qualitative comparison difference between score slope and Tweedie difference.}
Using the score slope introduces noticeable artifacts.}
    \label{fig:temporal_diff}
\end{minipage}
\hfill
% ===================== RIGHT COLUMN =====================
\begin{minipage}[t]{0.56\linewidth}
    \vspace{0pt}
    \centering

    % ---------- Top: figure ----------
    \includegraphics[width=\linewidth]{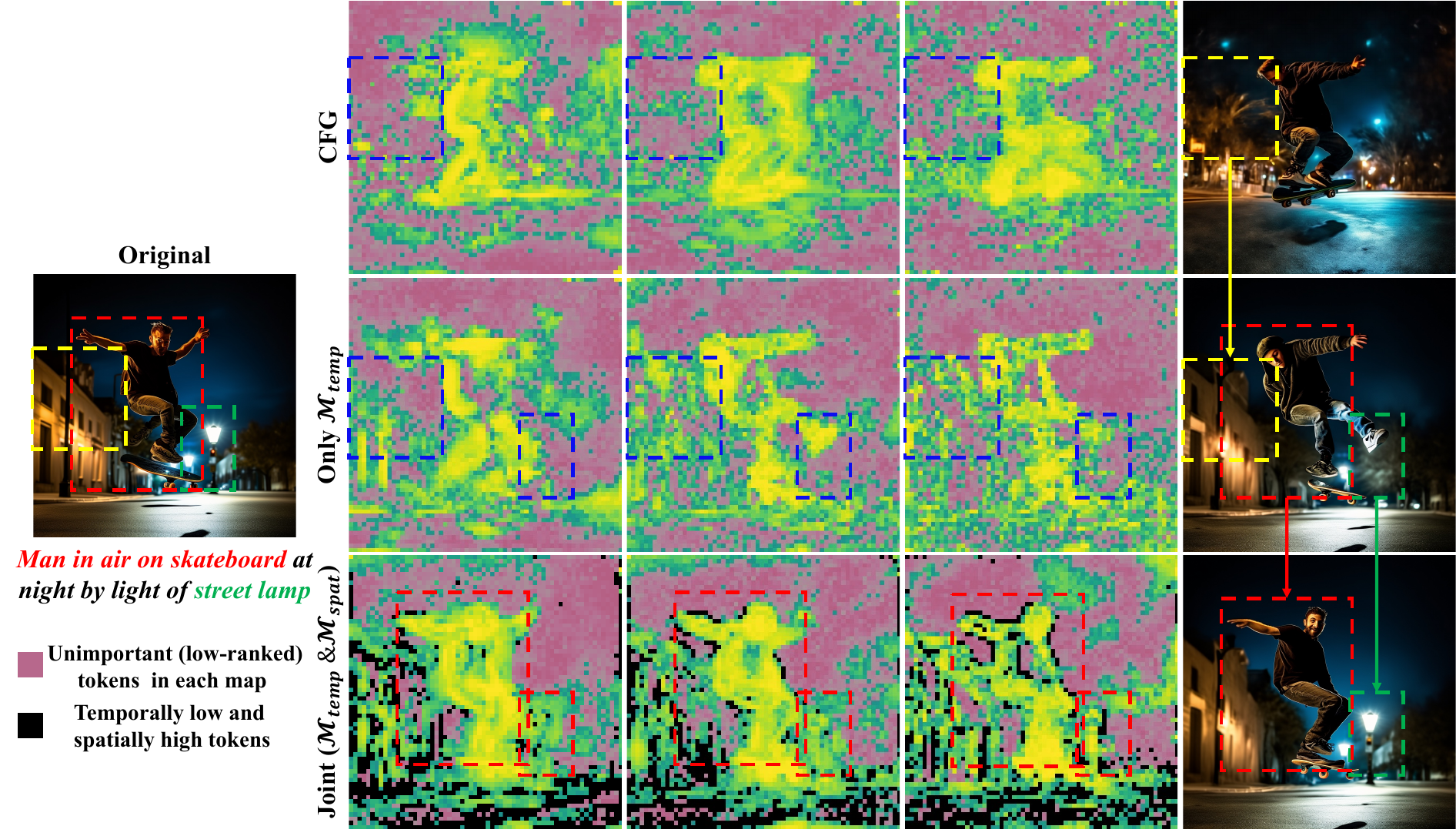}
    \vspace{2pt}
    \captionof{figure}{\textbf{Effect of different importance signals.}
    Jointly using temporal and spatial signals, the proposed importance map preserves both structurally important object regions and unconditioned background content.}
    \label{fig:signal_effect}

    \vspace{8pt}

    % ---------- Bottom: table ----------
    \captionof{table}{\textbf{Ablation on constraint-set definition $\mathcal{S}_{\mathrm{con}}$.}
    We compare different choices for selecting constraint tokens.}
    \label{tab:con_ablation}
    \vspace{2pt}

    \small
    \setlength{\tabcolsep}{4.2pt}
    \renewcommand{\arraystretch}{1.05}
    \resizebox{\linewidth}{!}{
    \begin{tabular}{l c c c c c}
    \toprule
     \textbf{Setting} & \textbf{CLIP-T}$\uparrow$ & \textbf{SSIM}$\uparrow$ & \textbf{PSNR}$\uparrow$ & \textbf{LPIPS}$\downarrow$ & \textbf{FID}$\downarrow$ \\
    \midrule
    None & \textbf{31.1419} & 0.8037 & 20.3736 & 0.1601 & 43.7692 \\
    $\mathcal{M}_{\mathrm{temp}}\uparrow \cap \:\mathcal{M}_{\mathrm{spat}}\uparrow$ & \underline{31.1375} & \textbf{0.8112} & \textbf{20.5136} & \textbf{0.1501} & \textbf{42.4474} \\
    $\mathcal{M}_{\mathrm{temp}}\downarrow \cap      \:\mathcal{M}_{\mathrm{spat}}\uparrow$ & 31.1372 & \underline{0.8076} & \underline{20.5027} & \underline{0.1569} & \underline{43.6605} \\
    $\mathcal{M}_{\mathrm{temp}}\uparrow \cap \:\mathcal{M}_{\mathrm{spat}}\downarrow$ & 31.1250 & 0.8005 & 20.3459 & 0.1641 & 43.8062 \\
    \bottomrule
    \end{tabular}
    }
\end{minipage}

\end{figure}

We conduct various ablation studies on PixArt-$\alpha$. Additional experiments are in Appendix~\ref{sec:add_exp}.

\noindent\textbf{Importance of Temporal and Spatial Signals.}
In Tab.~\ref{tab:axis_ablation}, we ablate the two components of our importance map by using only the temporal axis, only the spatial axis, or both.
Using both axes yields the best overall performance, confirming that temporal evolution and structural saliency provide complementary cues for token selection, whereas either axis alone is insufficient to fully characterize mergeability.
We further replace our temporal signal with the raw score slope to evaluate its effectiveness in the actual merging process.
As shown in Tab.~\ref{tab:axis_ablation}, this substitution consistently degrades all evaluation metrics, and Fig.~\ref{fig:temporal_diff} shows that it also introduces noticeable visual artifacts.
This is consistent with our earlier analysis that the score slope becomes increasingly contaminated by schedule-driven effects in later denoising steps, making it less reliable for token merging.

Fig.~\ref{fig:signal_effect} further illustrates how different importance signals affect region preservation.
The CFG-based signal~\cite{Wu_2025_ICCV} is strongly biased toward prompt-aligned foreground regions and therefore neglects unmentioned background content.
Using only the temporal signal alleviates this issue by preserving evolving background regions (yellow box), but it still misses structurally important boundaries (black regions), leading to structural distortions (red box).
By jointly modeling temporal and spatial signals, our importance map preserves both evolving background content and critical object boundaries, resulting in more faithful image structure and higher overall visual quality.

\noindent\textbf{Constraint Token Selection.}
In Tab.~\ref{tab:con_ablation}, we ablate different definitions of the constraint set.
The best results are obtained when the constraint tokens are selected from regions with both high temporal and high spatial responses.
In particular, including spatially high-response tokens improves structural preservation, indicating that the spatial axis captures fine object details and structurally salient regions.
By contrast, using spatially low-response tokens degrades performance because such tokens are biased toward homogeneous regions and therefore fail to protect meaningful object structure.

\begin{figure*}[t]
\centering

% =========================================================
% Left: Importance Signal Table
% =========================================================
\begin{minipage}[t]{0.51\textwidth}
\vspace{0pt}
\centering

\captionof{table}{
\textbf{Effectiveness of the proposed importance signal on PixArt-$\alpha$ at $r=0.3$.}
``Merging'' indicates the merging strategy adopted from either~\cite{Wu_2025_ICCV} or ours, while ``Signal'' indicates whether token importance is determined by CFG~\cite{Wu_2025_ICCV} or our proposed importance map.
}
\label{tab:importance_signal}

\vspace{2pt}

\small
\setlength{\tabcolsep}{3.2pt}
\renewcommand{\arraystretch}{1.08}

\resizebox{\linewidth}{!}{
\begin{tabular}{llccccc}
\toprule
\textbf{Merging}
& \textbf{Signal}
& \textbf{CLIP-T}$\uparrow$
& \textbf{HPSv2}$\uparrow$
& \textbf{SSIM}$\uparrow$
& \textbf{PSNR}$\uparrow$
& \textbf{LPIPS}$\downarrow$ \\
\midrule

CFG~\cite{Wu_2025_ICCV}
& CFG~\cite{Wu_2025_ICCV}
& 30.8801
& 27.7375
& 0.7137
& 17.8157
& 0.2369 \\

CFG~\cite{Wu_2025_ICCV}
& Ours
& \underline{31.0545}
& \underline{27.9366}
& \underline{0.7644}
& \underline{20.0917}
& \underline{0.1915} \\

Ours
& Ours
& \textbf{31.1375}
& \textbf{28.5591}
& \textbf{0.8112}
& \textbf{20.5136}
& \textbf{0.1501} \\

\bottomrule
\end{tabular}
}

\end{minipage}
\hfill
% =========================================================
% Right: Human Study Figure
% =========================================================
\begin{minipage}[t]{0.47\textwidth}
\vspace{0pt}
\centering

\includegraphics[width=\linewidth]{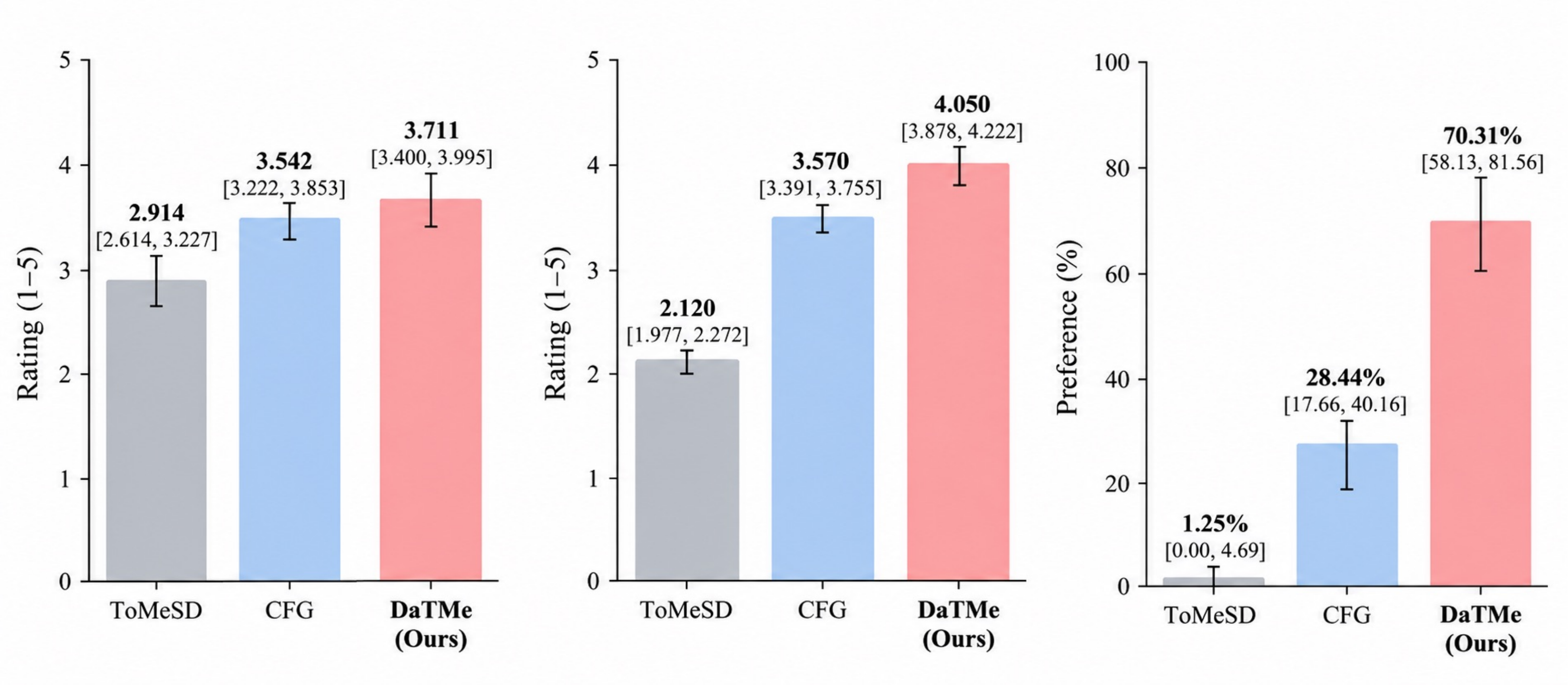}

\captionof{figure}{
\textbf{Human preference study on PixArt-$\alpha$ at $r=0.3$.}
DaTMe receives the highest ratings for text alignment and visual quality,
as well as the highest overall preference.
}
\label{fig:human_study}

\end{minipage}

\end{figure*}

\noindent\textbf{Effectiveness of the Importance Signal.}
To isolate the contribution of our importance signal, we keep the entire merging pipeline of~\cite{Wu_2025_ICCV} unchanged and replace only its CFG-based importance with ours.
Since~\cite{Wu_2025_ICCV} requires a single token ranking, we combine our temporal and spatial rankings using rank-sum.
As shown in Tab.~\ref{tab:importance_signal}, replacing only the CFG-based signal with ours improves CLIP-T, HPSv2, and all fidelity metrics, demonstrating that the proposed dynamics-aware importance more effectively identifies tokens to preserve or merge.
Using the full DaTMe merging strategy further improves all metrics, indicating that the role-assignment and spatial merging design provides additional complementary benefits.

\noindent\textbf{Preference Study.}
We evaluate perceptual quality at a compression ratio of $r=0.3$ using human judgments.
For 20 randomly sampled PixArt-$\alpha$ cases, 32 evaluators compare outputs from ToMeSD, CFG~\cite{Wu_2025_ICCV}, and DaTMe without access to the uncompressed reference.
Evaluators independently rate text alignment and visual quality, and select their overall preferred result.
As shown in Fig.~\ref{fig:human_study}, DaTMe achieves the highest ratings for both text alignment and visual quality, while also being the most preferred method overall.

\section{Conclusion}

We presented \textbf{DaTMe}, a diffusion-dynamics-aware token merging 
framework for efficient image generation.
By leveraging the Tweedie denoised estimate, DaTMe defines token 
importance along two complementary axes---temporal evolution and 
structural saliency---that reflect the image generation process.
This allows DaTMe to identify safely mergeable tokens while 
preserving structurally critical regions, achieving strong image fidelity and favorable
quality-efficiency trade-offs across DiT-based
diffusion models.
Our work shows that grounding token merging in the underlying 
generation dynamics is a key step toward faithful and efficient 
diffusion inference.

\begin{ack}
This work was partly supported by the Institute of Information \&
Communications Technology Planning \& Evaluation (IITP) grant funded by the
Korean government (MSIT) (No. RS-2025-25422680, Metacognitive AGI Framework)
and the Institute of Information \& Communications Technology Planning \&
Evaluation (IITP) grant funded by the Korean government (MSIT)
(No.RS-2025-02215122, Development and Demonstration of Lightweight AI Model
for Smart Homes).
% Use unnumbered first level headings for the acknowledgments. All acknowledgments
% go at the end of the paper before the list of references. Moreover, you are required to declare
% funding (financial activities supporting the submitted work) and competing interests (related financial activities outside the submitted work).
% More information about this disclosure can be found at: \url{https://neurips.cc/Conferences/2025/PaperInformation/FundingDisclosure}.

% Do {\bf not} include this section in the anonymized submission, only in the final paper. You can use the \texttt{ack} environment provided in the style file to automatically hide this section in the anonymized submission.
\end{ack}

\bibliographystyle{naturemag}
\bibliography{ref} 

\newpage
\appendix

\section{Formal Decomposition of Score Slope Instability}
\label{sec:formula}
We provide a formal analysis of why the score slope fails to reliably 
capture per-token convergence in late denoising steps.

\noindent\textbf{Decomposition of score slope.}
Recall that the score function is estimated as:
\begin{equation}
    s_t \approx -\frac{1}{\sigma_t}\epsilon_\theta(x_t, t, c).
\end{equation}
To understand the temporal variation of the score, we analyze the 
score slope between consecutive timesteps. For brevity, let 
$\epsilon_\theta(t)$ denote $\epsilon_\theta(x_t, t, c)$. 
The score slope is:
\begin{equation}
    s_t - s_{t-1} = \frac{\epsilon_\theta(t-1)}{\sigma_{t-1}} - \frac{\epsilon_\theta(t)}{\sigma_t}.
\end{equation}
By adding and subtracting the term $\frac{\epsilon_\theta(t)}{\sigma_{t-1}}$, 
\begin{equation}
    s_t - s_{t-1} 
    = \frac{\epsilon_\theta(t-1)}{\sigma_{t-1}} - \frac{\epsilon_\theta(t)}{\sigma_{t-1}} 
    + \frac{\epsilon_\theta(t)}{\sigma_{t-1}} - \frac{\epsilon_\theta(t)}{\sigma_t},
\end{equation}
we can decouple this slope into two distinct components:
\begin{equation}
    s_t - s_{t-1} 
    = \underbrace{\frac{\epsilon_\theta(t-1) - \epsilon_\theta(t)}{\sigma_{t-1}}}_{\text{difference term}} 
    + \underbrace{\epsilon_\theta(t) \left( \frac{1}{\sigma_{t-1}} - \frac{1}{\sigma_t} \right)}_{\text{scale term}}.
    \label{eq:score_decomp}
\end{equation}
This decomposition reveals a limitation. 
The first term captures the actual change in the 
model's noise prediction, reflecting semantic token-level evolution. 
However, the second term is entirely dictated by 
the predefined noise schedule $\sigma_t$, scaled by the current 
noise prediction $\epsilon_\theta(t)$.

\noindent\textbf{Behavior in late denoising steps.}
As generation progresses toward the final steps, the noise scale 
$\sigma_t$ becomes extremely small. Because $\sigma_t$ decays 
rapidly, the inverse difference $\left| \frac{1}{\sigma_{t-1}} - \frac{1}{\sigma_t} \right|$ 
grows disproportionately large. 
Meanwhile, since $\epsilon_\theta$ is trained to match samples 
from $\mathcal{N}(0,I)$~\cite{ho2020denoising}, its norm is 
concentrated near $\sqrt{D}$ in expectation, where $D$ is 
the dimensionality~\cite{karras2022elucidating}.
Consequently, the scale term in Eq.~\eqref{eq:score_decomp} 
significantly amplifies the raw noise prediction $\epsilon_\theta(t)$, 
eventually overshadowing the difference term. 
Therefore, in later denoising stages, the score slope $s_t - s_{t-1}$ 
is heavily dominated by the noise schedule scaling rather than 
actual changes in each token, which is consistent with the high autocorrelation 
and loss of discriminative power observed in Fig.~\ref{fig:autocorr}.

\noindent\textbf{Decomposition of Tweedie difference.}
To understand why the Tweedie difference avoids this scale-induced amplification, we first define the Tweedie denoised estimate. 
Under variance-preserving diffusion (VP diffusion)~\cite{song2020score, ho2020denoising}, the noisy state is 
$x_t = \alpha_t x_0 + \sigma_t \epsilon$, where the signal and 
noise schedules satisfy $\alpha_t^2 + \sigma_t^2 = 1$. 
The posterior mean estimate of the clean data is~\cite{efron2011tweedie}:
\begin{equation}
    \hat{x}_0(t) = \mathbb{E}[x_0 \mid x_t] \approx \frac{x_t - \sigma_t \epsilon_\theta(t)}{\alpha_t},
\end{equation}
where the approximation arises from using the learned $\epsilon_\theta(t)$ 
in place of the true noise.
This clean-space projection is embedded within the deterministic 
update rule (e.g., DDIM~\cite{song2020denoising}). The latent 
transition from timestep $t$ to $t-1$ is:
\begin{equation}
    x_{t-1} = \alpha_{t-1} \underbrace{\left( \frac{x_t - \sigma_t \epsilon_\theta(t)}{\alpha_t} \right)}_{\text{Tweedie estimate } \hat{x}_0(t)} + \sigma_{t-1} \epsilon_\theta(t).
\end{equation}
Since the parenthesized term is exactly $\hat{x}_0(t)$, we can write:
\begin{equation}
    x_{t-1} = \alpha_{t-1} \hat{x}_0(t) + \sigma_{t-1} \epsilon_\theta(t).
\end{equation}
We decompose the inter-step difference by substituting this 
$x_{t-1}$ into the Tweedie estimate at the next timestep, 
$\hat{x}_0(t-1) = \frac{x_{t-1} - \sigma_{t-1} \epsilon_\theta(t-1)}{\alpha_{t-1}}$:
\begin{equation}
    \hat{x}_0(t-1) 
    = \frac{\big(\alpha_{t-1} \hat{x}_0(t) + \sigma_{t-1} \epsilon_\theta(t)\big) - \sigma_{t-1} \epsilon_\theta(t-1)}{\alpha_{t-1}}.
\end{equation}
Simplifying yields:
\begin{equation}
    \hat{x}_0(t-1) = \hat{x}_0(t) + \frac{\sigma_{t-1}}{\alpha_{t-1}} \big(\epsilon_\theta(t) - \epsilon_\theta(t-1)\big).
\end{equation}
Rearranging, the Tweedie inter-step difference reduces to:
\begin{equation}
    \hat{x}_0(t) - \hat{x}_0(t-1) = \frac{\sigma_{t-1}}{\alpha_{t-1}} \big(\epsilon_\theta(t-1) - \epsilon_\theta(t)\big).
    \label{eq:tweedie_decomp}
\end{equation}

Comparing Eq.~\eqref{eq:tweedie_decomp} with Eq.~\eqref{eq:score_decomp} 
reveals a difference in how the noise schedule enters each formulation.
First, the problematic scale term---the $1/\sigma_t$ 
scaling that overshadowed the genuine signal--- no longer dominates the slope. The Tweedie difference cleanly isolates the 
difference term change, $\epsilon_\theta(t-1) - \epsilon_\theta(t)$.
Second, instead of amplifying this change with a rapidly growing 
$1/\sigma_{t-1}$ multiplier, the Tweedie difference scales it by 
$\frac{\sigma_{t-1}}{\alpha_{t-1}}$ (closely related to the inverse 
signal-to-noise ratio~\cite{luo2022understandingdiffusionmodelsunified}). 
As $t \to 0$, we have $\sigma_{t-1} \to 0$ and $\alpha_{t-1} \to 1$, 
so $\frac{\sigma_{t-1}}{\alpha_{t-1}} \to 0$. This decays 
the variation as the network's prediction stabilizes, ensuring that 
the Tweedie difference provides a stable metric for convergence 
without scale-induced amplification.

\noindent\textbf{Extension to flow matching.}
The Tweedie formulation extends naturally to flow matching 
architectures such as FLUX~\cite{flux2024}. While VP diffusion 
satisfies $\alpha_t^2 + \sigma_t^2 = 1$, FLUX adopts the 
\emph{rectified flow} schedule $\alpha_t = 1 - t,\ \sigma_t = t$, 
which does not satisfy the VP constraint. Both nonetheless share 
the general form $x_t = \alpha_t x_0 + \sigma_t \epsilon$, allowing 
the Tweedie derivation to proceed with FLUX-specific schedule 
values. Since rectified flow uses continuous time $t \in [0, 1]$, 
we denote the step size between consecutive sampling times by 
$\Delta t$.

\textbf{Clean-image estimate under velocity parameterization.}
Unlike VP models that predict noise $\epsilon_\theta$, rectified 
flow models predict the velocity field 
$v_\theta(t) \approx \epsilon - x_0$~\cite{flux2024}. From the 
forward process $x_t = (1-t) x_0 + t\,\epsilon$ and the velocity 
definition $\epsilon = v_\theta(t) + x_0$, we have:
\begin{equation}
    x_t = (1-t) x_0 + t\big(v_\theta(t) + x_0\big) = x_0 + t \cdot v_\theta(t).
\end{equation}
Solving for $x_0$ yields the FLUX clean-image estimate:
\begin{equation}
    \hat{x}_0(t) = x_t - t \cdot v_\theta(t),
    \label{eq:flux_tweedie}
\end{equation}
which is the velocity-parameterization analog of 
Eq.~\eqref{eq:tweedie}.

\textbf{Tweedie inter-step difference under Euler update.}
For rectified flow, the deterministic Euler step from time $t$ to 
$t - \Delta t$ is:
\begin{equation}
    x_{t-\Delta t} = x_t - \Delta t \cdot v_\theta(t).
    \label{eq:flux_euler}
\end{equation}
Applying Eq.~\eqref{eq:flux_tweedie} at $t-\Delta t$ and 
substituting Eq.~\eqref{eq:flux_euler}:
\begin{align}
    \hat{x}_0(t-\Delta t) 
    &= x_{t-\Delta t} - (t-\Delta t) \cdot v_\theta(t-\Delta t) \nonumber\\
    &= \big[x_t - \Delta t \cdot v_\theta(t)\big] - (t-\Delta t) \cdot v_\theta(t-\Delta t).
\end{align}
Subtracting from $\hat{x}_0(t) = x_t - t \cdot v_\theta(t)$:
\begin{align}
    \hat{x}_0(t) - \hat{x}_0(t-\Delta t) 
    &= -t \cdot v_\theta(t) + \Delta t \cdot v_\theta(t) + (t-\Delta t) \cdot v_\theta(t-\Delta t) \nonumber\\
    &= -(t-\Delta t) \cdot v_\theta(t) + (t-\Delta t) \cdot v_\theta(t-\Delta t) \nonumber\\
    &= (t-\Delta t)\big(v_\theta(t-\Delta t) - v_\theta(t)\big).
    \label{eq:tweedie_flux}
\end{align}

Eq.~\eqref{eq:tweedie_flux} mirrors Eq.~\eqref{eq:tweedie_decomp}: 
the Tweedie inter-step difference isolates the change in 
prediction—velocity rather than noise—scaled by the schedule 
factor $(t-\Delta t)$, which decays to zero as $t \to 0$. The 
Tweedie difference therefore remains a stable temporal signal 
under flow matching as well.

\section{DaTMe Details}
\label{sec:datme_detail}
\noindent\textbf{DaTMe Process.}
In Alg.~\ref{alg:datme_step}, we show the detailed step-by-step process of our proposed DaTMe framework.

\begin{algorithm}[t]
\caption{Dynamics-Aware Token Merging (DaTMe) at timestep $t$}
\label{alg:datme_step}
\begin{algorithmic}[1]

\Require token features $\mathbf{F}^{(t)}$,
temporal map $\mathcal{M}_{\mathrm{temp}}^{(t)}$,
spatial map $\mathcal{M}_{\mathrm{spat}}^{(t)}$,
compression ratio $r$, number of grids $P$

\Ensure merged token features $\mathbf{F}^{\prime(t)}$

\State $N \gets H \times W$,
\quad $\mathcal{S}_{\mathrm{tot}} \gets \{1,\ldots,N\}$,
\quad $n_{\mathrm{tgt}} \gets \lfloor rN \rfloor$

\Statex
\State \textbf{// 1. Dynamics-Aware Role Assignment}

\State $\mathcal{S}_{\mathrm{con}}
\gets
\mathrm{TopFrac}_{1-r}
\bigl(\mathcal{M}_{\mathrm{temp}}^{(t)}\bigr)
\cap
\mathrm{TopFrac}_{1-r}
\bigl(\mathcal{M}_{\mathrm{spat}}^{(t)}\bigr)$
\Comment{protected tokens}

\State $\mathcal{S}_{\mathrm{src}}
\gets
\mathrm{BotFrac}_{r}
\bigl(\mathcal{M}_{\mathrm{temp}}^{(t)}\bigr)
\cap
\mathrm{BotFrac}_{r}
\bigl(\mathcal{M}_{\mathrm{spat}}^{(t)}\bigr)$
\Comment{safe merge sources}

\Statex
\State \textbf{// 2. Spatially Distributed Source Completion}

\State $\mathcal{U}
\gets
\mathcal{S}_{\mathrm{tot}}
\setminus
\bigl(
\mathcal{S}_{\mathrm{con}}
\cup
\mathcal{S}_{\mathrm{src}}
\bigr)$

\State $\mathcal{S}_{\mathrm{src}}^{\mathrm{add}}
\gets
\mathrm{GridSample}
\Bigl(
\mathcal{U},
n_{\mathrm{tgt}}-|\mathcal{S}_{\mathrm{src}}|,
P
\Bigr)$

\State $\mathcal{S}_{\mathrm{src}}^{*}
\gets
\mathcal{S}_{\mathrm{src}}
\cup
\mathcal{S}_{\mathrm{src}}^{\mathrm{add}}$

\Statex
\State \textbf{// 3. Locality-Constrained Spatial Matching}

\State $\mathcal{D}
\gets
\mathcal{S}_{\mathrm{tot}}
\setminus
\bigl(
\mathcal{S}_{\mathrm{con}}
\cup
\mathcal{S}_{\mathrm{src}}^{*}
\bigr)$
\Comment{destination tokens}

\For{each $j \in \mathcal{S}_{\mathrm{src}}^{*}$}
    \State $d(j)
    \gets
    \arg\min_{k\in\mathcal{D}}
    \|\mathbf{p}_j-\mathbf{p}_k\|_2$
\EndFor

\State $\mathbf{F}^{\prime(t)}
\gets
\mathrm{Merge}
\Bigl(
\mathbf{F}^{(t)},
\{j \rightarrow d(j)\}_{j\in\mathcal{S}_{\mathrm{src}}^{*}}
\Bigr)$
\Comment{$\mathcal{S}_{\mathrm{con}}$ remains unchanged}

\State \Return $\mathbf{F}^{\prime(t)}$

\end{algorithmic}
\end{algorithm}

\noindent\textbf{Additional Implementation Details.}
For PixArt-$\alpha$, we use 20 inference steps with a guidance scale of 4.5. Token merging is applied to transformer layers 9–19 starting from the second denoising step, since the temporal signal requires the previous Tweedie estimate, which is unavailable at the first step. For FLUX, images are generated using 35 inference steps with a guidance scale of 3.5. Within the FLUX architecture, we apply token merging to blocks 9 through 19 out of the 19 dual-stream blocks, and to blocks 7 through 22 out of the 38 single-stream blocks.

We further evaluate DaTMe on U-Net-based architectures, specifically Stable Diffusion 1.5 and SDXL. For these models, we set the interpolation scale to $s=2$ and the number of grid cells to $P=4$. Images are generated using 50 inference steps, with a guidance scale of 7.5 for Stable Diffusion 1.5 and 5.0 for SDXL.
Following~\cite{Wu_2025_ICCV, bolya2023token}, we apply token merging at the $64\times 64$ latent resolution in both architectures.
Unlike in DiT-based models, we do not explicitly construct the constraint set $\mathcal{S}_{\mathrm{con}}$ for U-Net backbones. Because U-Net architectures already possess a strong spatial inductive bias through convolutions, tokens with similar features are naturally organized in nearby spatial regions~\cite{d2021convit}. Consequently, spatial locality is implicitly preserved. We therefore only identify the source set $\mathcal{S}_{\mathrm{src}}$ and sample the remaining merge candidates without enforcing an additional constraint set. Finally, token merging begins at the same denoising step as in the DiT experiments.
% Unlike in DiT-based models, we do not explicitly construct the constraint set $\mathcal{S}_{\mathrm{con}}$ for U-Net backbones.
% This is because U-Net architectures already possess strong spatial inductive bias through convolution, so tokens with similar features are naturally organized in nearby spatial regions~\cite{d2021convit}.
% As a result, locality is already implicitly preserved, and we only identify $\mathcal{S}_{\mathrm{src}}$ while sampling the remaining merge candidates without an additional constraint set.
% Finally, we start merging from the same denoising step as in the DiT-based setting.
% \sjcha{FLUX, etc...}

\section{Additional Experiments}
\label{sec:add_exp}
\begin{figure}[h]
    \centering
    \includegraphics[width=0.98\linewidth]{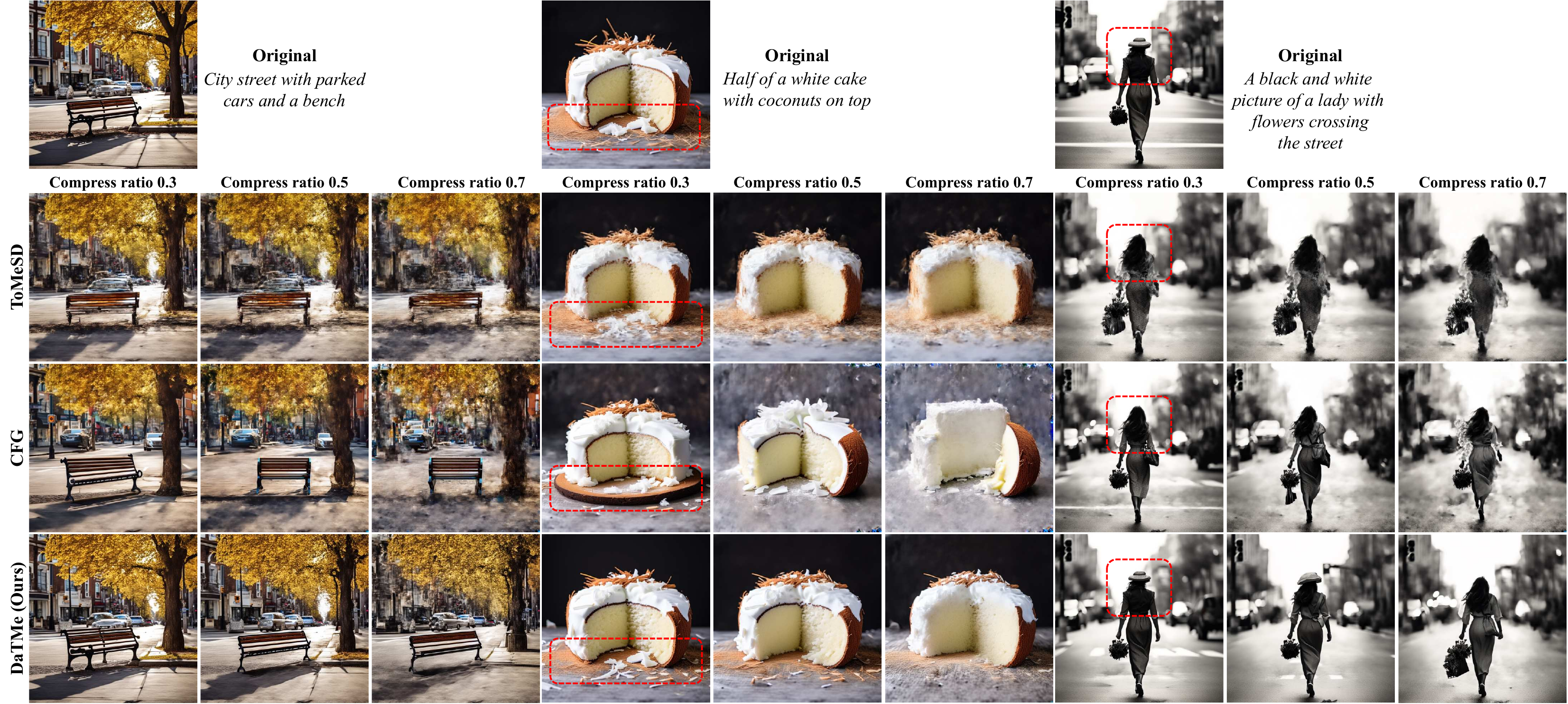}
\caption{\textbf{Additional PixArt-$\alpha$ results under different compression ratios.} Unlike prior methods, DaTMe more faithfully preserves the fine details present in the original image.
}
    \label{fig:pixart_supple}
\end{figure}

\begin{figure}[h]
    \centering
    \includegraphics[width=0.98\linewidth]{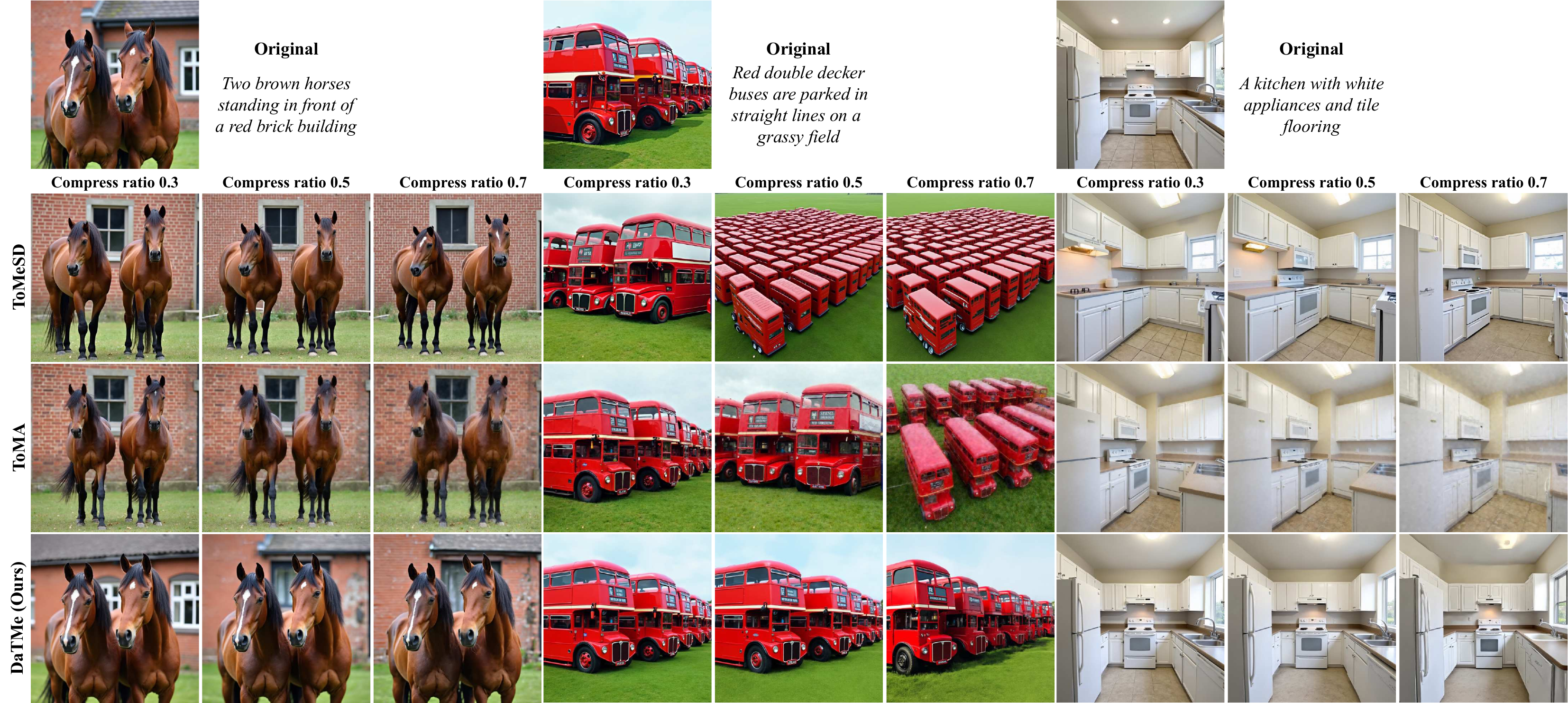}
\caption{\textbf{Additional FLUX results under different compression ratios.} Unlike prior methods, DaTMe more faithfully preserves the fine details present in the original image.
}
    \label{fig:flux_supple}
\end{figure}

\noindent\textbf{Additional Comparison Results.}
We compare additional results on PixArt-$\alpha$ under different compression ratios in Fig.~\ref{fig:pixart_supple}.
While the CFG-based method degrades noticeably at higher compression ratios, DaTMe more faithfully preserves subject details and unconditioned content, such as the woman's hat and the floor decoration beneath the cake (red boxes).
% In FLUX, while ToMeSD method shows quite similar image compared to the original image, the structure of the image dramatically changes at higher compression ratios compared with DaTMe (Fig.~\ref{fig:flux_supple}).
On FLUX, existing methods increasingly deviate from the
original generation as compression increases (Fig.~\ref{fig:flux_supple}).
ToMeSD introduces changes in spatial composition and
object structure, while ToMA additionally exhibits noisy
visual degradation.
In contrast, DaTMe more faithfully preserves the original
composition and fine details across compression ratios.

\begin{table}[t]
\centering

\begin{minipage}[t]{0.60\linewidth}
\centering
\caption{\textbf{Text alignment and efficiency score on SD 1.5.}}
\label{tab:sd15_main_metrics}
\vspace{2pt}

\small
\setlength{\tabcolsep}{3.5pt}
\renewcommand{\arraystretch}{1.05}
\resizebox{\linewidth}{!}{
\begin{tabular}{l c c c c c}
\toprule
& \textbf{CLIP-T}$\uparrow$ & \textbf{HPSv2}$\uparrow$ & \textbf{FID}$\downarrow$ & \textbf{Latency (s)}$\downarrow$ & \textbf{TFLOPs}$\downarrow$ \\
\midrule
SD 1.5 (Base) & 31.2669 & 25.5643 & 33.2610 & 1.0186 & 70.396 \\
\midrule
\multicolumn{6}{c}{\textbf{Compression ratio 0.3}} \\
\midrule
ToMeSD & 31.3262 & 25.1050 & 32.7718 & 1.0283 & 70.744 \\
CFG & \underline{31.3453} & \underline{25.1062} & \textbf{32.6889} & 1.0669 & 70.744 \\
DaTMe (Ours) & \textbf{31.3456} & \textbf{25.1185} & 33.3034 & 0.9155 & 70.849 \\
\midrule
\multicolumn{6}{c}{\textbf{Compression ratio 0.5}} \\
\midrule
ToMeSD & 31.3212 & 24.6347 & \textbf{33.4531} & 1.0141 & 70.408 \\
CFG & \textbf{31.3631} & \underline{24.6633} & \underline{33.4947} & 1.0704 & 70.408 \\
DaTMe (Ours) & \underline{31.3306} & \textbf{24.6682} & 34.3004 & 0.9138 & 70.727 \\
\midrule
\multicolumn{6}{c}{\textbf{Compression ratio 0.7}} \\
\midrule
ToMeSD & 31.2225 & 23.6237 & 36.3672 & 1.0028 & 70.073 \\
CFG & \underline{31.2227} & \textbf{23.7960} & \textbf{34.9584} & 1.0507 & 70.073 \\
DaTMe (Ours) & \textbf{31.2550} & \underline{23.7728} & \underline{35.7699} & 0.9025 & 70.192 \\
\bottomrule
\end{tabular}
}
\end{minipage}
\hfill
\begin{minipage}[t]{0.38\linewidth}
\centering
\caption{\textbf{Image fidelity score on SD 1.5.}}
\label{tab:sd15_fidelity_metrics}
\vspace{2pt}

\small
\setlength{\tabcolsep}{4pt}
\renewcommand{\arraystretch}{1.05}
\resizebox{\linewidth}{!}{
\begin{tabular}{l c c c}
\toprule
& \textbf{SSIM}$\uparrow$ & \textbf{PSNR}$\uparrow$ & \textbf{LPIPS}$\downarrow$ \\
\midrule
\multicolumn{4}{c}{\textbf{Compression ratio 0.3}} \\
\midrule
ToMeSD & 0.8245 & 20.0186 & \underline{0.1435} \\
CFG & \underline{0.8265} & \underline{20.1034} & 0.1448 \\
DaTMe (Ours) & \textbf{0.8579} & \textbf{21.2101} & \textbf{0.1171} \\
\midrule
\multicolumn{4}{c}{\textbf{Compression ratio 0.5}} \\
\midrule
ToMeSD & 0.7575 & 18.4332 & 0.2038 \\
CFG & \underline{0.7596} & \underline{18.4743} & \underline{0.2036} \\
DaTMe (Ours) & \textbf{0.7820} & \textbf{19.1236} & \textbf{0.1842} \\
\midrule
\multicolumn{4}{c}{\textbf{Compression ratio 0.7}} \\
\midrule
ToMeSD & \underline{0.7001} & \underline{17.2531} & \underline{0.2648} \\
CFG & 0.6804 & 17.1733 & 0.2730 \\
DaTMe (Ours) & \textbf{0.7035} & \textbf{17.5194} & \textbf{0.2611} \\
\bottomrule
\end{tabular}
}
\end{minipage}

\end{table}

\begin{table}[t]
\centering

\begin{minipage}[t]{0.60\linewidth}
\centering
\caption{\textbf{Text alignment and efficiency score on SDXL.}}
\label{tab:sdxl_main_metrics}
\vspace{2pt}

\small
\setlength{\tabcolsep}{3.5pt}
\renewcommand{\arraystretch}{1.05}
\resizebox{\linewidth}{!}{
\begin{tabular}{l c c c c c}
\toprule
& \textbf{CLIP-T}$\uparrow$ & \textbf{HPSv2}$\uparrow$ & \textbf{FID}$\downarrow$ & \textbf{Latency (s)}$\downarrow$ & \textbf{TFLOPs}$\downarrow$ \\
\midrule
SDXL (Base) & 31.9501 & 27.0768 & 32.5454 & 5.0761 & 607.849 \\
\midrule
\multicolumn{6}{c}{\textbf{Compression ratio 0.3}} \\
\midrule
ToMeSD & 31.9603 & \underline{26.7230} & 32.4035 & 5.0839 & 607.851 \\
CFG & \textbf{31.9795} & 26.7159 & \textbf{32.3833} & 5.2559 & 607.851 \\
DaTMe (Ours) & \underline{31.9622} & \textbf{26.9697} & \underline{32.3880} & 4.2480 & 608.035 \\
\midrule
\multicolumn{6}{c}{\textbf{Compression ratio 0.5}} \\
\midrule
ToMeSD & 31.9411 & 26.1445 & 33.3079 & 5.0216 & 605.164 \\
CFG & \textbf{32.0154} & \underline{26.1695} & \underline{32.8330} & 5.2054 & 605.164 \\
DaTMe (Ours) & \underline{31.9604} & \textbf{26.8040} & \textbf{32.4385} & 4.2622 & 607.312 \\
\midrule
\multicolumn{6}{c}{\textbf{Compression ratio 0.7}} \\
\midrule
ToMeSD & 31.6383 & 24.4532 & 39.8369 & 4.9837& 602.481 \\
CFG & \textbf{31.9262} & \underline{24.9657} & \underline{36.0965} & 5.1569 & 602.481 \\
DaTMe (Ours) & \underline{31.8824} & \textbf{25.1762} & \textbf{35.2577} & 4.2672 & 605.930 \\
\bottomrule
\end{tabular}
}
\end{minipage}
\hfill
\begin{minipage}[t]{0.38\linewidth}
\centering
\caption{\textbf{Image fidelity score on SDXL.}}
\label{tab:sdxl_fidelity_metrics}
\vspace{2pt}

\small
\setlength{\tabcolsep}{4pt}
\renewcommand{\arraystretch}{1.05}
\resizebox{\linewidth}{!}{
\begin{tabular}{l c c c}
\toprule
& \textbf{SSIM}$\uparrow$ & \textbf{PSNR}$\uparrow$ & \textbf{LPIPS}$\downarrow$ \\
\midrule
\multicolumn{4}{c}{\textbf{Compression ratio 0.3}} \\
\midrule
ToMeSD & 0.8076 & \underline{21.1286} & \underline{0.1715} \\
CFG & \underline{0.8078} & 21.1278 & \underline{0.1715} \\
DaTMe (Ours) & \textbf{0.8413} & \textbf{22.2688} & \textbf{0.1412} \\
\midrule
\multicolumn{4}{c}{\textbf{Compression ratio 0.5}} \\
\midrule
ToMeSD & \underline{0.7458} & \underline{19.5331} & \underline{0.2367} \\
CFG & 0.7408 & 19.3888 & 0.2433 \\
DaTMe (Ours) & \textbf{0.7801} & \textbf{20.3747} & \textbf{0.2007} \\
\midrule
\multicolumn{4}{c}{\textbf{Compression ratio 0.7}} \\
\midrule
ToMeSD & \underline{0.6942} & \underline{18.5884} & \underline{0.3097} \\
CFG & 0.6827 & 18.1728 & 0.3194 \\
DaTMe (Ours) & \textbf{0.7114} & \textbf{18.8536} & \textbf{0.2860} \\
\bottomrule
\end{tabular}
}
\end{minipage}

\end{table}

\noindent\textbf{Extended Experiments on U-Net Architecture.}
We further evaluate DaTMe on U-Net-based diffusion models under different compression ratios.
For Stable Diffusion 1.5, Tab.~\ref{tab:sd15_main_metrics} and Tab.~\ref{tab:sd15_fidelity_metrics} show that DaTMe better preserves structural fidelity while maintaining strong text alignment across compression ratios.
A similar trend is observed on SDXL in Tab.~\ref{tab:sdxl_main_metrics} and Tab.~\ref{tab:sdxl_fidelity_metrics}, demonstrating the effectiveness of DaTMe beyond transformer-based diffusion architectures.

\begin{figure}[t]
    \centering
    \includegraphics[width=0.98\linewidth]{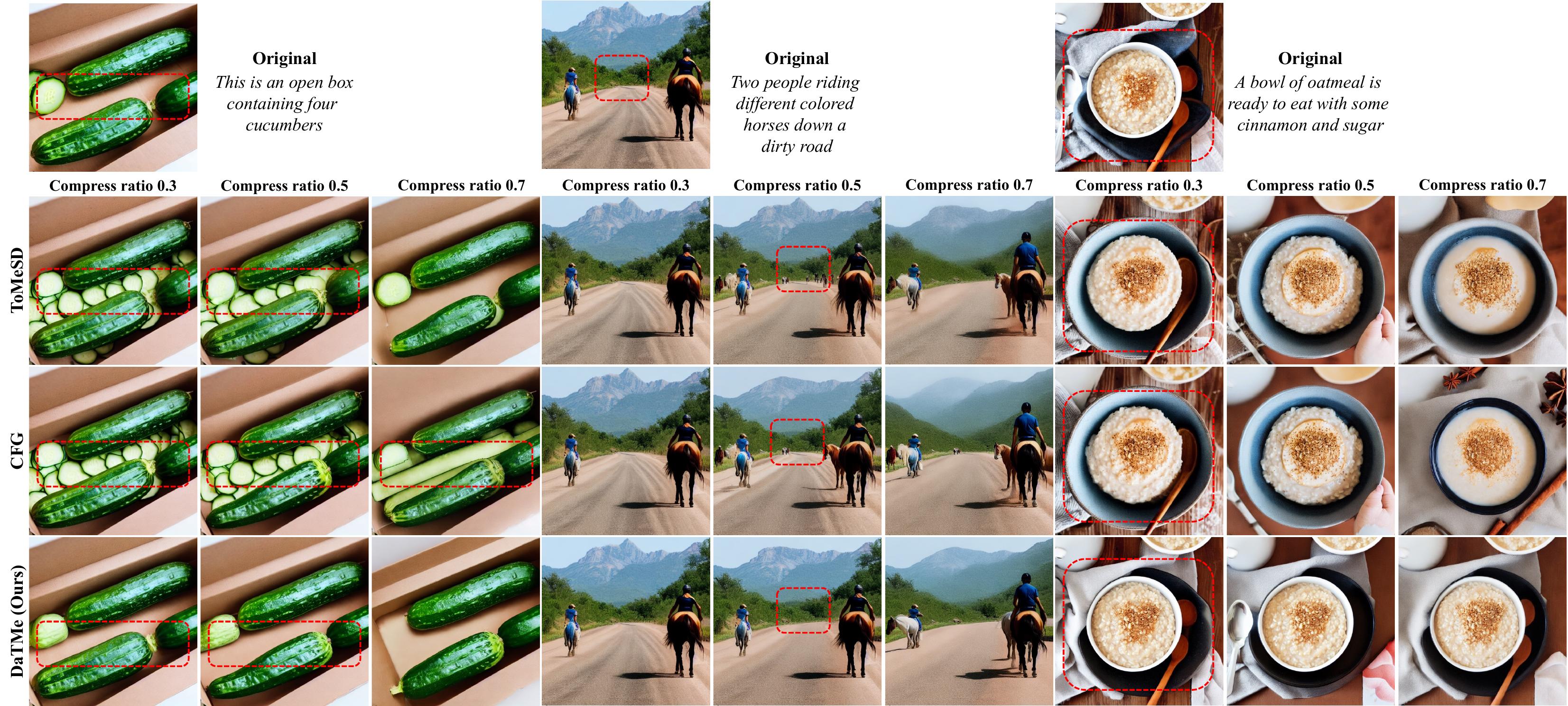}
\caption{\textbf{SD 1.5 results under different compression ratios.} Unlike prior methods, DaTMe more faithfully preserves the contents of the original image.
}
    \label{fig:sd_supple}
\end{figure}

\begin{figure}[t]
    \centering
    \includegraphics[width=0.98\linewidth]{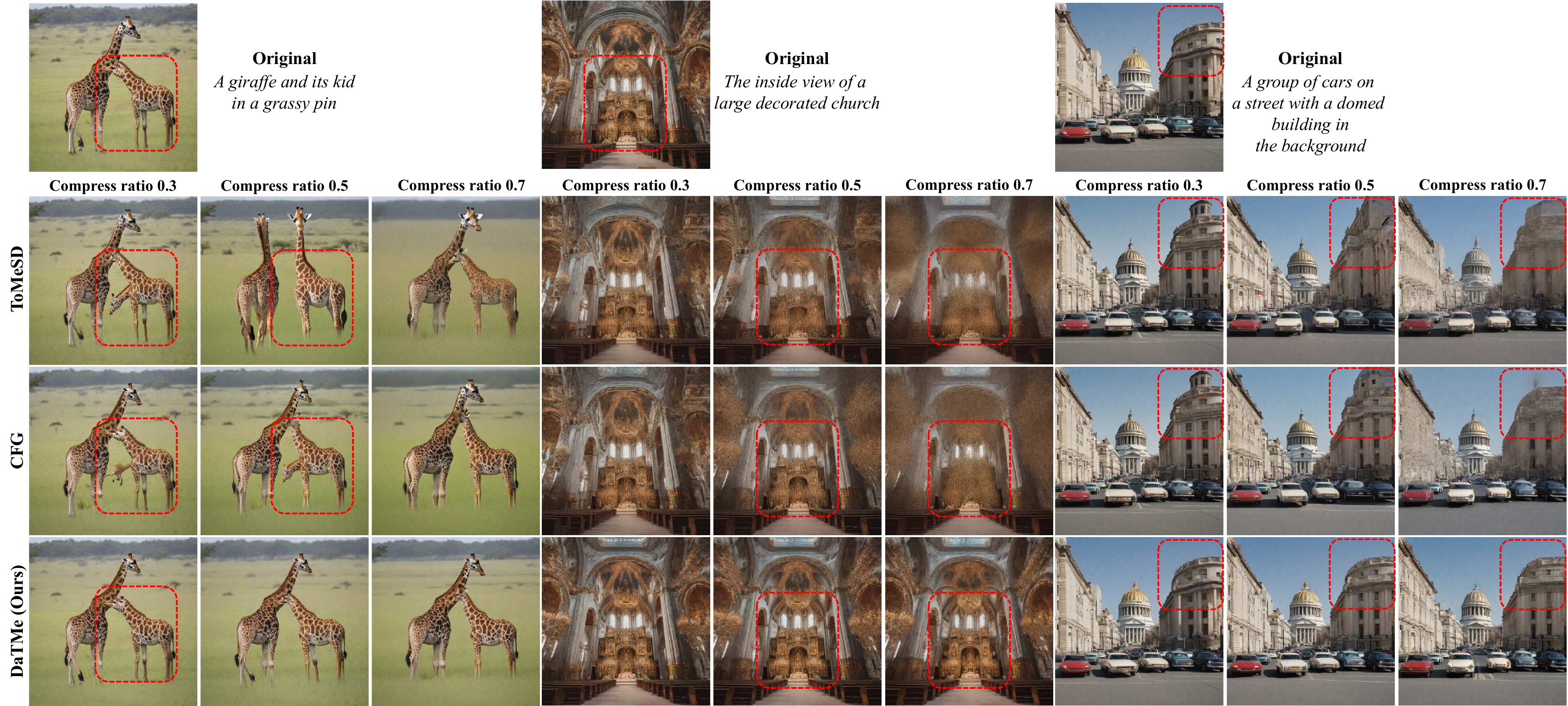}
\caption{\textbf{SDXL results under different compression ratios.} Unlike prior methods, DaTMe more faithfully preserves the contents of the original image especially background details.
}
    \label{fig:sdxl_supple}
\end{figure}

This is further supported by Fig.~\ref{fig:sd_supple} and Fig.~\ref{fig:sdxl_supple}: prior methods often fail to preserve the original image structure, particularly in non-prompt regions highlighted by red boxes, whereas DaTMe maintains these regions more faithfully.

\noindent\textbf{Additional Ablation Studies on PixArt-$\alpha$.}
Tab.~\ref{tab:grid_ablation} studies the number of spatial grid cells used for grid-based source completion.
Increasing the number of grid cells generally improves structure preservation by distributing additional source tokens more evenly across the image, but overly fine grids introduce extra latency.
We therefore choose $P=4$, which achieves the best overall quality while remaining faster than the unmerged PixArt-$\alpha$ baseline.
Fig.~\ref{fig:grid_ablation_vis} further visualizes this trend: denser grids better preserve structural details by preventing source tokens from concentrating in a limited region.
Tab.~\ref{tab:laplacian_ablation} ablates the down/up-sampling scale used to construct the spatial axis.
We vary the scale from $2$ to $16$ and set $s=4$, which achieves the best overall trade-off across quality metrics.

\begin{figure}[h]
    \centering
    \includegraphics[width=0.98\linewidth]{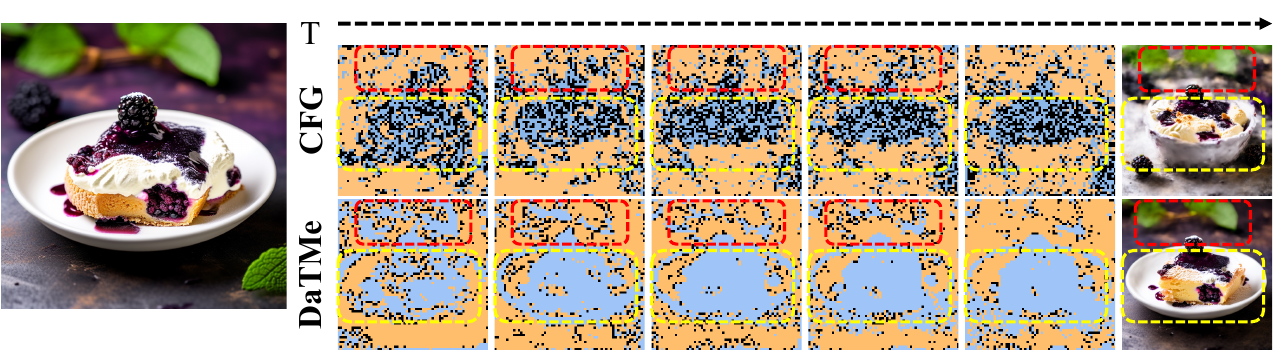}
\caption{\textbf{Difference in merging maps.} Black spots indicate the destinations of merged source tokens. Our method better preserves structurally important details across the entire image.}
    \label{fig:merge_map}
\end{figure}

Fig.~\ref{fig:merge_map} presents the merging maps over denoising timesteps. Our method better preserves structural details, such as the bowl shape and leaves, thereby yielding higher-fidelity images.

\noindent\textbf{Ablation Studies on SD.}
We select the hyperparameters for U-Net based model based on Tab.~\ref{tab:con_ablation_sd}, Tab.~\ref{tab:laplacian_ablation_sd} and Tab.~\ref{tab:grid_ablation_sd}.

\begin{figure}[t]
\centering

\begin{minipage}[t]{0.48\linewidth}
\centering
\includegraphics[width=\linewidth]{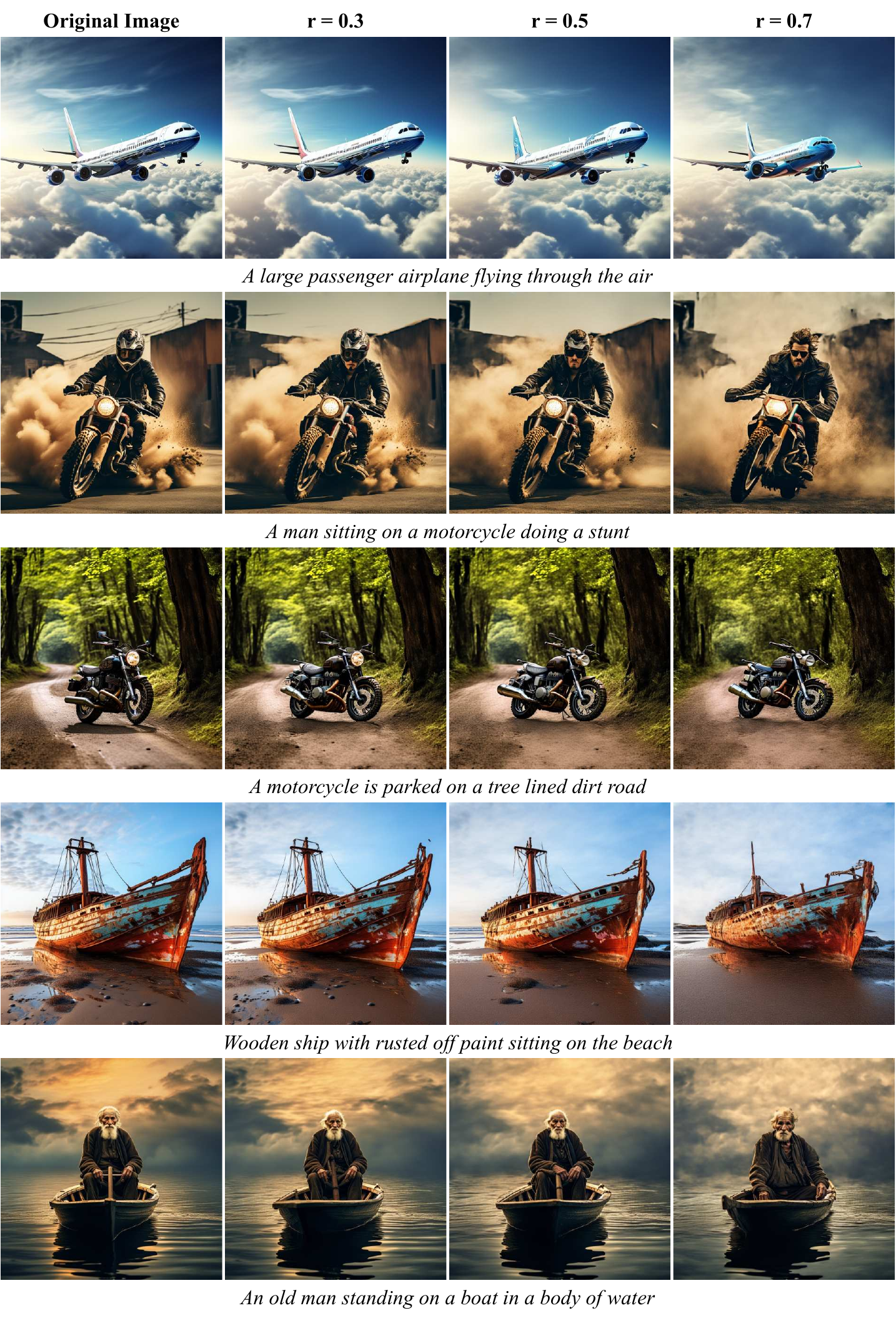}
\caption{\textbf{Additional PixArt-$\alpha$ results.}}
\label{fig:pixart_datme_additional}
\end{minipage}
\hfill
\begin{minipage}[t]{0.48\linewidth}
\centering
\includegraphics[width=\linewidth]{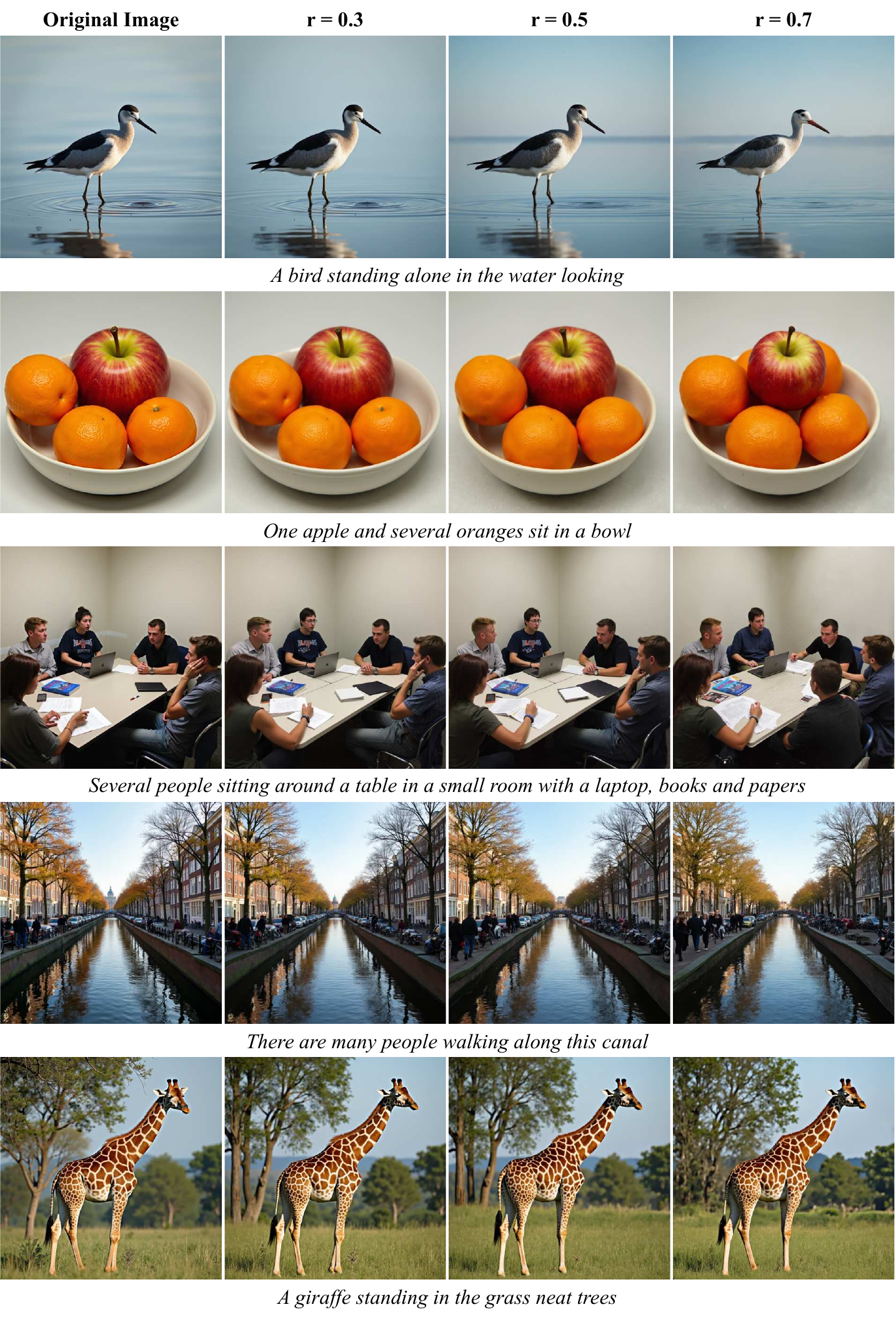}
\caption{\textbf{Additional FLUX results.}}
\label{fig:flux_datme_additional}
\end{minipage}

\end{figure}

\noindent\textbf{DaTMe Additional Results.}
In Fig.~\ref{fig:pixart_datme_additional} and Fig.~\ref{fig:flux_datme_additional}, we present additional samples under different compression ratios.
DaTMe consistently preserves the structural details of the original generation while maintaining strong text alignment.

\begin{table*}[t]
\centering

\begin{minipage}[t]{0.48\linewidth}
\centering
\small
\setlength{\tabcolsep}{4.2pt}
\renewcommand{\arraystretch}{1.08}
\caption{\textbf{Ablation on the number of spatial grids in PixArt-$\alpha$.}
We vary the number of grid cells used for grid-based source completion and select $P=4$ as the best trade-off between latency and performance.}
\label{tab:grid_ablation}
\resizebox{\linewidth}{!}{
\begin{tabular}{c c c c c c c}
\toprule
\textbf{Grid} & \textbf{CLIP-T}$\uparrow$ & \textbf{SSIM}$\uparrow$ & \textbf{PSNR}$\uparrow$ & \textbf{LPIPS}$\downarrow$ & \textbf{FID}$\downarrow$ & \textbf{Latency (s)}$\downarrow$ \\
\midrule
PixArt-$\alpha$~\cite{chen2023pixartalphafasttrainingdiffusion} & 31.2485 & -- & -- & -- & 42.1001 & 1.9417 \\
CFG~\cite{Wu_2025_ICCV} & 30.8801 & 0.7137 & 17.8157 & 0.2369 & 44.6070 & 1.7458 \\ \hline
1  & \textbf{31.1426} & 0.8099 & 20.4707 & 0.1513 & 42.5585 & \textbf{1.7343}{\scriptsize\textcolor{blue}{$_{-0.2074}$}} \\
4  & \underline{31.1375} & 0.8112 & 20.5136 & 0.1501 & 42.4474 & 1.8030{\scriptsize\textcolor{blue}{$_{-0.1387}$}} \\
16 & 31.1374 & \underline{0.8128} & \underline{20.5705} & \underline{0.1485} & \textbf{42.3615} & 1.9765{\scriptsize\textcolor{red}{$_{+0.0348}$}} \\
64 & 31.1307 & \textbf{0.8140} & \textbf{20.6223} & \textbf{0.1477} & \underline{42.3792} & 2.4443{\scriptsize\textcolor{red}{$_{+0.5026}$}} \\
\bottomrule
\end{tabular}
}
\end{minipage}
\hfill
\begin{minipage}[t]{0.48\linewidth}
\centering
\small
\setlength{\tabcolsep}{4.2pt}
\renewcommand{\arraystretch}{1.08}
\caption{\textbf{Ablation on the interpolation scale.}
We vary the down/up-sampling scale used to construct the spatial axis in PixArt-$\alpha$.}
\label{tab:laplacian_ablation}
\resizebox{\linewidth}{!}{
\begin{tabular}{c c c c c c}
\toprule
\textbf{Scale} & \textbf{CLIP-T}$\uparrow$ & \textbf{SSIM}$\uparrow$ & \textbf{PSNR}$\uparrow$ & \textbf{LPIPS}$\downarrow$ & \textbf{FID}$\downarrow$ \\
\midrule
2  & 31.1368 & \underline{0.8102} & \underline{20.4854} & \underline{0.1507} & \underline{42.5226} \\
4  & \underline{31.1375} & \textbf{0.8112} & \textbf{20.5136} & \textbf{0.1501} & \textbf{42.4474} \\
8  & \textbf{31.1445} & 0.8091 & 20.4785 & 0.1522 & 42.6547 \\
16 & 31.1259 & 0.8016 & 20.2553 & 0.1596 & 42.7489 \\
\bottomrule
\end{tabular}
}
\end{minipage}

\end{table*}

\begin{figure}[t]
\centering

\begin{minipage}[t]{0.48\linewidth}
\vspace{0pt}
\centering
\includegraphics[width=\linewidth]{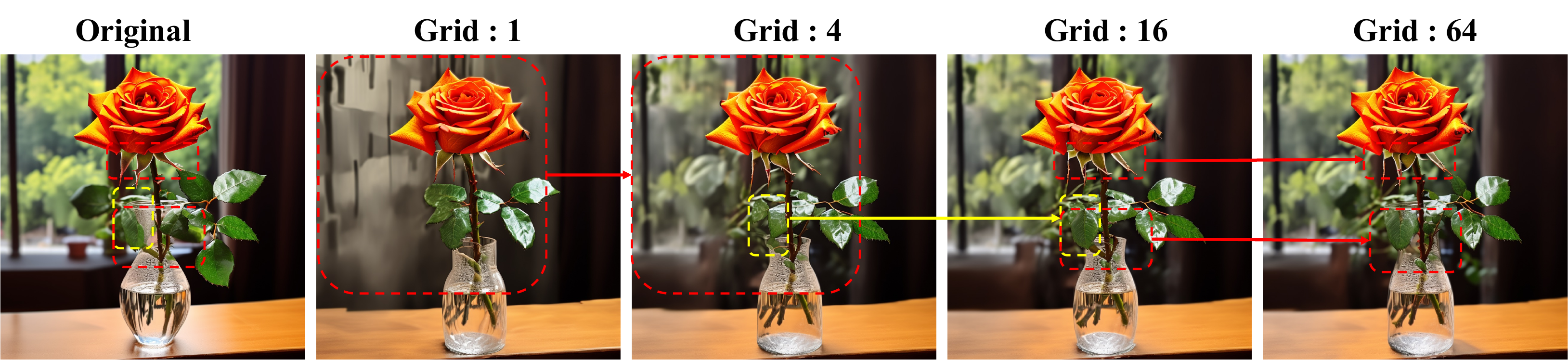}
\caption{\textbf{Visualization of different grid sizes on PixArt-$\alpha$.}
Denser grids better preserve structural details by preventing source tokens from concentrating in a limited region.}
\label{fig:grid_ablation_vis}
\end{minipage}
\hfill
\begin{minipage}[t]{0.48\linewidth}
\vspace{0pt}
\centering
\captionof{table}{\textbf{Ablation on constraint token selection in SD 1.5.}}
\label{tab:con_ablation_sd}
\vspace{2pt}

\small
\setlength{\tabcolsep}{4.2pt}
\renewcommand{\arraystretch}{1.05}
\resizebox{\linewidth}{!}{
\begin{tabular}{l c c c c c}
\toprule
 & \textbf{CLIP-T}$\uparrow$ & \textbf{SSIM}$\uparrow$ & \textbf{PSNR}$\uparrow$ & \textbf{LPIPS}$\downarrow$ & \textbf{FID}$\downarrow$ \\
\midrule
$\mathcal{M}_{\mathrm{temp}}\uparrow \cap \mathcal{M}_{\mathrm{spat}}\downarrow$ 
& 31.3346 & 0.8402 & 20.6266 & 0.1316 & \textbf{33.1811} \\
$\mathcal{M}_{\mathrm{temp}}\downarrow \cap \mathcal{M}_{\mathrm{spat}}\uparrow$ 
& \textbf{31.3592} & \underline{0.8529} & 21.0306 & \underline{0.1214} & 33.4103 \\
$\mathcal{M}_{\mathrm{temp}}\uparrow \cap \mathcal{M}_{\mathrm{spat}}\uparrow$ 
& \underline{31.3527} & 0.8472 & \underline{20.8985} & 0.1273 & 33.3237 \\
None 
& 31.3456 & \textbf{0.8579} & \textbf{21.2101} & \textbf{0.1171} & \underline{33.3034} \\
\bottomrule
\end{tabular}
}
\end{minipage}

\end{figure}

\begin{table*}[t]
\centering

\begin{minipage}[t]{0.48\linewidth}
\centering
\small
\setlength{\tabcolsep}{5.0pt}
\renewcommand{\arraystretch}{1.08}
\caption{\textbf{Ablation on the interpolation scale in SD 1.5.}}
\label{tab:laplacian_ablation_sd}
\resizebox{\linewidth}{!}{
\begin{tabular}{c c c c c c}
\toprule
\textbf{Scale} & \textbf{CLIP-T}$\uparrow$ & \textbf{SSIM}$\uparrow$ & \textbf{PSNR}$\uparrow$ & \textbf{LPIPS}$\downarrow$ & \textbf{FID}$\downarrow$ \\
\midrule
2  & 31.3456 & \textbf{0.8579} & \textbf{21.2101} & \textbf{0.1171} & \textbf{33.3034} \\
4  & \textbf{31.3543} & \underline{0.8569} & \underline{21.1812} & \underline{0.1182} & 33.4557 \\
8  & \underline{31.3468} & 0.8541 & 21.1044 & 0.1209 & \underline{33.4293} \\
16 & 31.3422 & 0.8514 & 21.0544 & 0.1237 & 33.4424 \\
\bottomrule
\end{tabular}
}
\end{minipage}
\hfill
\begin{minipage}[t]{0.48\linewidth}
\centering
\small
\setlength{\tabcolsep}{5.0pt}
\renewcommand{\arraystretch}{1.08}
\caption{\textbf{Ablation on the number of spatial grids in SD 1.5.}}
\label{tab:grid_ablation_sd}
\resizebox{\linewidth}{!}{
\begin{tabular}{c c c c c c}
\toprule
\textbf{Grid} & \textbf{CLIP-T}$\uparrow$ & \textbf{SSIM}$\uparrow$ & \textbf{PSNR}$\uparrow$ & \textbf{LPIPS}$\downarrow$ & \textbf{FID}$\downarrow$ \\
\midrule
1  & \textbf{31.3529} & 0.8571 & 21.1678 & 0.1178 & 33.3703 \\
4  & \underline{31.3456} & \textbf{0.8579} & \textbf{21.2101} & \textbf{0.1171} & \textbf{33.3034} \\
16 & 31.3443 & \underline{0.8576} & \underline{21.1988} & \underline{0.1176} & \underline{33.3631} \\
64 & 31.3365 & 0.8570 & 21.1814 & 0.1180 & 33.3841 \\
\bottomrule
\end{tabular}
}
\end{minipage}

\end{table*}

% \clearpage
\section{Limitations}
Similar to prior token merging methods, DaTMe faces limitations under aggressive token compression.
\begin{wrapfigure}[14]{r}{0.45\linewidth}
\vspace{-8pt}
\centering
\includegraphics[width=\linewidth]{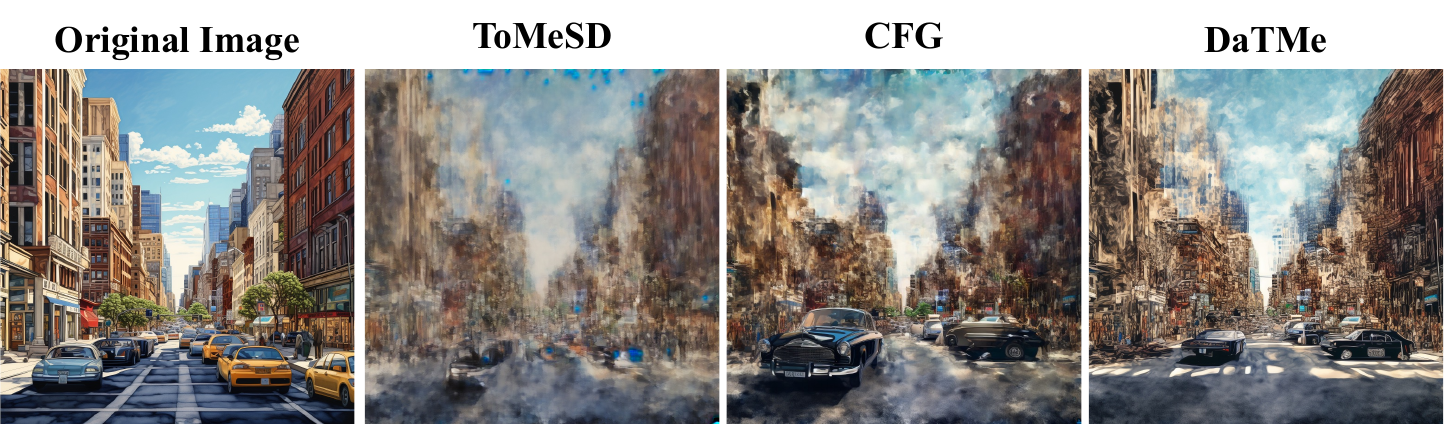}
\caption{\textbf{Limitations.}
In highly complex images, aggressive compression adversely affects image quality, leading to loss of structural details and visual coherence.}
\label{fig:your_label}
\vspace{-10pt}
\end{wrapfigure}
In highly complex scenes with many objects or intricate backgrounds, excessive merging can inevitably remove necessary tokens, leading to degraded generation quality.
In addition, the layer-wise roles of diffusion models remain underexplored, especially in DiT-based architectures.
Understanding which layers contribute most to structural layout, fine details, or background refinement could further improve token merging by enabling layer-aware compression strategies.

\begin{figure}[t]
    \centering
    \includegraphics[width=0.98\linewidth]{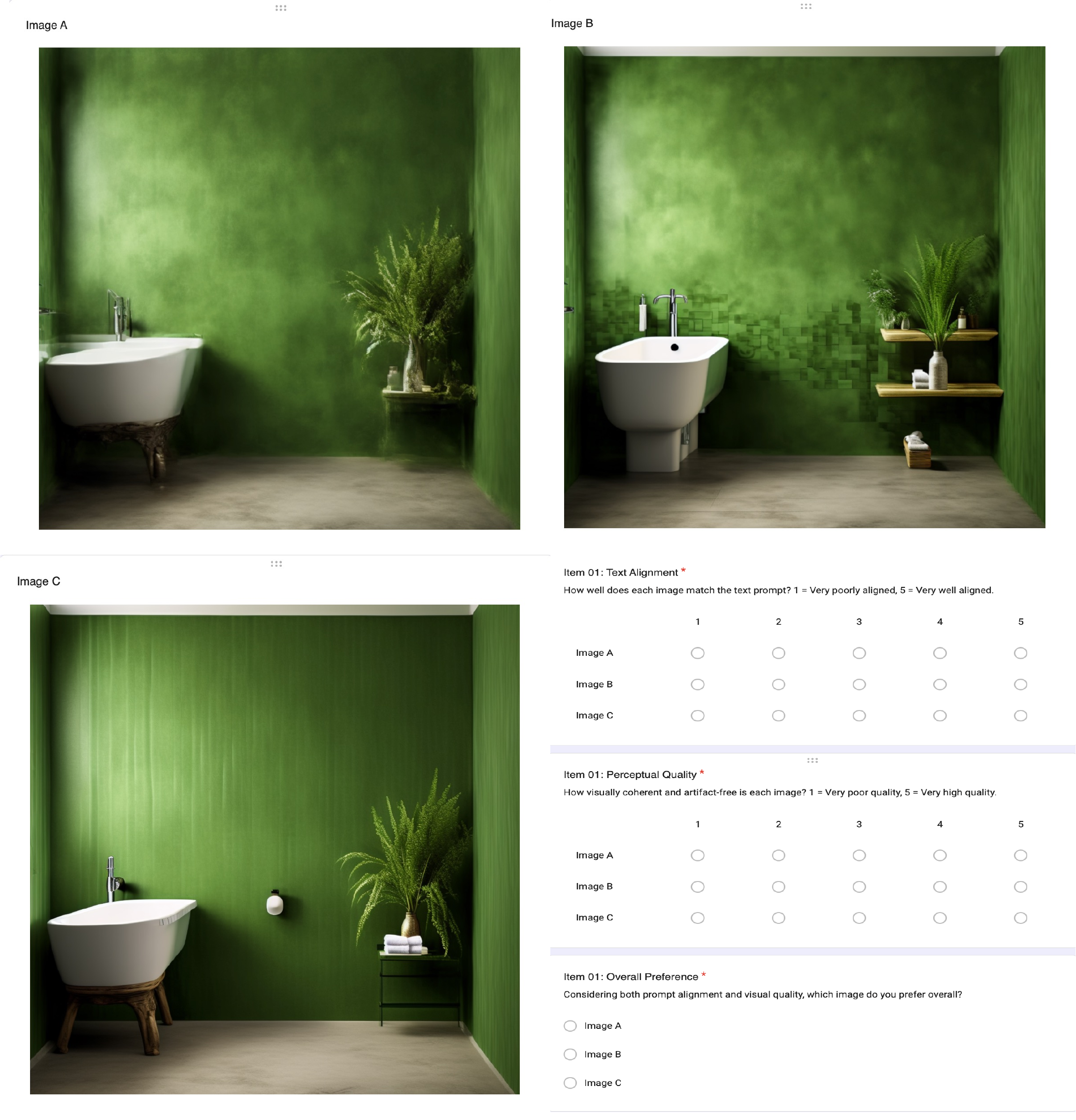}
\caption{\textbf{Example questionnaire for the human preference study.}
Each evaluation sample includes three questions.}
    \label{fig:humanstudy_example}
\end{figure}

%%%%%%%%%%%%%%%%%%%%%%%%%%%%%%%%%%%%%%%%%%%%%%%%%%%%%%%%%%%%

%%%%%%%%%%%%%%%%%%%%%%%%%%%%%%%%%%%%%%%%%%%%%%%%%%%%%%%%%%%%

% \newpage
% \input{checklist.tex}

\end{document}